\documentclass[sigconf,screen,nonacm]{acmart} % <-- this is for the preprint version
\AtBeginDocument{%
  \providecommand\BibTeX{{%
    \normalfont B\kern-0.5em{\scshape i\kern-0.25em b}\kern-0.8em\TeX}}}

\copyrightyear{2021}
\acmYear{2021}
\setcopyright{acmlicensed}
\acmConference[CSCS '21]{Computer Science in Cars Symposium}{November 30, 2021}{Ingolstadt, Germany}
\acmBooktitle{Computer Science in Cars Symposium (CSCS '21), November 30, 2021, Ingolstadt, Germany}
\acmPrice{15.00}
\acmDOI{10.1145/3488904.3493376}
\acmISBN{978-1-4503-9139-9/21/11}

\usepackage{graphicx}
\usepackage{amsmath}
\usepackage{xspace}
\usepackage{array}
\usepackage{algorithm}% http://ctan.org/pkg/algorithm
\usepackage{algpseudocode}% http://ctan.org/pkg/algorithmicx
\newcolumntype{P}[1]{>{\centering\arraybackslash}p{#1}}
\usepackage{algorithm,algpseudocode}
\usepackage{xcolor}
\usepackage{hhline}
\usepackage{multirow}
\usepackage{subcaption}

\usepackage[font=small]{caption}
\usepackage[sort&compress,capitalize,noabbrev,nameinlink]{cleveref}
\creflabelformat{equation}{#2#1#3}
\Crefname{equation}{Eq.}{Eqs.}

\makeatletter
\DeclareRobustCommand\onedot{\futurelet\@let@token\@onedot}
\def\@onedot{\ifx\@let@token.\else.\null\fi\xspace}

\def\eg{\emph{e.g}\onedot} 
\def\ie{\emph{i.e}\onedot} 
\def\cf{\emph{c.f}\onedot} 
\def\etc{\emph{etc}\onedot} 
\def\wrt{w.r.t\onedot} 
\def\etal{\emph{et al}\onedot}
\makeatother

\begin{document}

\title{Multi-scale Iterative Residuals for Fast and Scalable Stereo Matching}

\author{Kumail Raza}
\email{kumail.raza@dfki.de}
\author{René Schuster}
\email{rene.schuster@dfki.de}
\author{Didier Stricker}
\email{didier.stricker@dfki.de}
\affiliation{%
  \institution{German Research Center for Artificial Intelligence -- DFKI}
  \city{Kaiserslautern}
  \country{Germany}
}
% \author{René Schuster}
% \email{rene.schuster@dfki.de}
% \affiliation{%
%   \institution{German Research Center for Artificial Intelligence -- DFKI}
%   \city{Kaiserslautern}
%   \country{Germany}
% }
% \author{Didier Stricker}
% \email{didier.stricker@dfki.de}
% \affiliation{%
%   \institution{German Research Center for Artificial Intelligence -- DFKI}
%   \city{Kaiserslautern}
%   \country{Germany}
% }

%\renewcommand{\shortauthors}{Raza \etal}

\begin{abstract}
    Despite the remarkable progress of deep learning in stereo matching, there exists a gap in accuracy between real-time models and slower state-of-the-art models which are suitable for practical applications. This paper presents an iterative multi-scale coarse-to-fine refinement (iCFR) framework to bridge this gap by allowing it to adopt any stereo matching network to make it fast, more efficient and scalable while keeping comparable accuracy. To reduce the computational cost of matching, we use multi-scale warped features to estimate disparity residuals and push the disparity search range in the cost volume to a minimum limit. Finally, we apply a refinement network to recover the loss of precision which is inherent in multi-scale approaches. We test our iCFR framework by adopting the matching networks from state-of-the art GANet and AANet. The result is 49$\times$ faster inference time compared to GANet-deep and 4$\times$ less memory consumption, with comparable error. Our best performing network, which we call FRSNet is scalable even up to an input resolution of 6K on a GTX 1080Ti, with inference time still below one second and comparable accuracy to AANet+. It out-performs all real-time stereo methods and achieves competitive accuracy on the KITTI benchmark.
\end{abstract}




\maketitle

%%%%%%%%% BODY TEXT
\section{Introduction}
Stereo matching is one of the top research areas in modern computer vision, with enormous applications in the industry particularly in autonomous driving and robotics. The idea is to reconstruct dense 3D geometry by estimating the disparity between image pixels in a rectified stereo image pair. To this end, dense matching pixel correspondences are used. Since, this is one of the major classical problems in computer vision, it has been studied substantially for more than half a century  \cite{survey} and has been matured. A huge amount of literature is available proposing various architectures both classical and modern, which aim to solve the problem of stereo matching successfully. A large part of these are deep neural networks. Although most of these networks provide sub-pixel accuracy with a state-of-the-art pixel outlier rate, a small number of them concentrate on the execution time and model growth, especially in the case of end-to-end trainable networks. The architectures with a design focus on run-time have much worse end point errors (EPE) and pixel error rates (ER) and are not comparable to the state-of-the-art models at all \cite{stereonet,deeppruner,ganet}.

\par The problem of stereo matching is solved traditionally with some key steps namely: 1) Extracting features, 2) Cost volume construction, 3) Cost aggregation for matching, 4) Regressing the final disparity. Two types of approaches exist in the literature incorporating some or all of these steps. These are direct disparity regression without building a cost volume and by using cost volume filtering. The first category disregards geometric variations by calculating disparity on dense matching pixel correspondences. These approaches although being relatively fast lead to large EPE and ER  \cite{sgm}. The latter namely \textit{volumetric methods}, construct and refine a cost volume. The idea is to build a higher dimensional feature volume containing all the candidate disparity values up to a maximum disparity range. These approaches are slower in classical processing as they incorporate some customized versions of dynamic programming algorithms for finding the best matching features over the cost volume. Neural networks built on volumetric matching often use two sub-networks in an end-to-end architecture. The first sub-net extracts features from the stereo image pairs which are used to build a 3D or 4D cost volume. The second network then uses 3D convolutions to compute matching costs in the cost volume before finally regressing the final disparity values. The run-time for these models is also nowhere comparable to real-time as these have to employ heavy operations on the high dimensional cost volume.
\par There exists a gap in the literature between real-time models with higher EPE and ER and slower state-of-the-art models with lower EPE and ER. We aim to bridge this gap by proposing a novel multi-scale coarse-to-fine refinement framework to estimate disparity by leveraging the already produced features usually at different resolutions by the feature extraction backbone and using much shallower cost volumes. These lower resolution disparity maps are then iteratively refined using the notion of disparity residuals. The idea being that these lower resolution operations on smaller disparity search ranges will be computationally more efficient and much more scalable in terms of memory while keeping the EPE and ER comparable. As shown in \cref{fig:growth} and \cref{tab:performance_comparison_kitti}, our approach bridges the accuracy gap between the real-time and state-of-the-art stereo matching models such as GANet \cite{ganet} and AANet \cite{aanet}, while staying as close as possible to real-time performance.
\par We introduce the notion of a \textit{prediction head}, which can be adopted from any stereo matching architecture and is used in our iterative coarse-to-fine refinement algorithm (iCFR). Separate heads are produced for each scale and their parameters are learned in an end-to-end fashion. We also propose a final refinement layer to recover the loss of precision that is inherent in down and upsampling operations. The resulting final network which we call FRSNet (Fast iterative Residual Stereo Network), using the iCFR, outperforms all available real-time architectures for stereo matching such as StereoNet \cite{stereonet}, DispNet \cite{dispnet}, Toast \cite{toast} and other deeper networks like GC-Net \cite{gcnet} while producing comparable EPE and ER to modern state-of-the-art networks such as AANet \cite{aanet}, PSMNet \cite{psmnet} and GANet \cite{ganet}, on the Sceneflow dataset \cite{dispnet} and the KITTI benchmark \cite{kitti}.
\begin{figure*}[t!]
    \includegraphics[width=\textwidth]{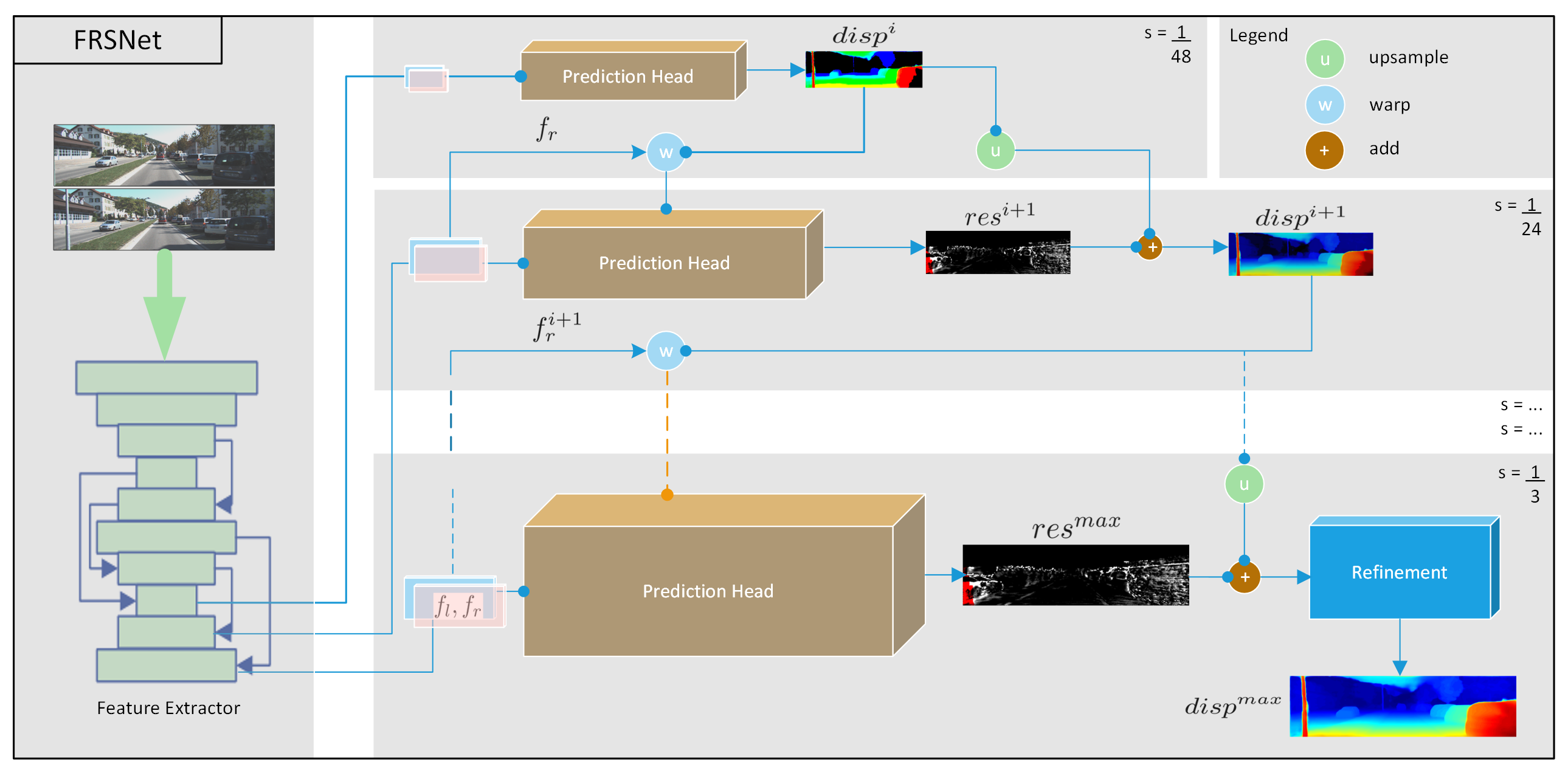}
    \caption{Overview of the architecture of our proposed FRSNet. Given a rectified stereo image pair, a stacked hourglass network with residual connections extracts the features. As the decoder in the stacked hourglass produces features at different scales, we employ them directly in the proposed iCFR algorithm, instead of down-sampling them. In the first iteration, the smallest resolution features are used \ie at $\frac{1}{48}$ of the original resolution to calculate a dense but very coarse disparity map $disp^{i}$. This full disparity map is then used to warp the target image features $f_{r}^{i+1}$ of the next scale \ie $\frac{1}{24}$. Using these warped features, the prediction head produces a residual disparity map $res^{i+1}$ this time. This is then added to the upsampled $disp^{i'}$ to get $disp^{i+1}$. The process is continued for all the remaining scales until passing the final disparity $disp^{i+4}$ through the refinement block to get the final disparity map $disp^{max}$ at input resolution.}
    \label{fig:frsnet}
\end{figure*}

\section{Related Work}
Traditional algorithms for stereo matching \cite{sgm, blockstereo, fuseproposals, nonlocalcost, compredundancy} yield discontinuous disparity maps with higher pixel outlier rates. This is because they usually use feature-based matching which does not take the geometric variations in images into account. Thus, the matching is ambiguous, due to occlusions, reflections, texture-less surfaces and repetitive patterns. Volumetric methods with cost volume construction and cost aggregation strategies were developed to achieve better matching \cite{nonlocalcost,segmentation,continuouspatch}.
Then with the advent of deep learning, these strategies transformed into learnable layers in terms of convolution and other operations. The first deep learning architecture for stereo matching was proposed in MC-CNN \cite{mccnn}, which used a feature extraction network instead of handcrafted features. Then, DispNet \cite{dispnet} provided a real-time end-to-end disparity estimation approach by using an encoder-decoder to directly regress the final disparity map. Since this network does not use prior information, it is completely data-driven and requires a huge dataset to train on. To this end, the authors of DispNet also presented the synthetic Sceneflow dataset \cite{dispnet} to train such kind of end-to-end networks, which we also partially use to train our network. Then a two-stage convolutional neural network was presented by Pang \etal \cite{cascaderesidual}, which separates the workflow for feature extraction and disparity refinement. The first approach with a 4D cost volume was proposed in GC-Net \cite{gcnet}. It applies 3D convolutions to aggregate the matching costs. GC-Net presents a three stage end-to-end trainable network for stereo estimation following the key steps of stereo matching \ie feature extraction, cost aggregation and disparity regression. It achieved state-of-the-art accuracy, however the inference times are much higher compared to the real-time models.
\par PSMNet \cite{psmnet} uses a feature pyramid network and stacked hourglass network along with twenty-five convolution layers to achieve state-of-the-art EPE and ER. However, higher number of convolutional layers deteriorate the run-time quite considerably and puts it far from the real-time category. GANet \cite{ganet} proposes to incorporate geometric optimization algorithms like SGM \cite{sgm} in deep learning pipelines as learnable layers. To this end, it presents the concept of guided aggregation layers in the end-to-end pipeline. These layers aim to leverage the local neighbourhood information as well as the global image context into guiding convolutions in the cost aggregation network. These layers push the sub-pixel accuracy even further and replace a large number of computationally complex 3D convolutions. LEAStereo \cite{leastereo} applies neural architecture search to choose the best possible parameters for a set of operations performed in the network for stereo matching. It achieves state-of-the-art accuracy on KITTI benchmark as well as close to real-time performance. This however, requires enormous amounts of computational resources to compute the best neural architecture for the job. To cater for this, the authors applied task-specific human knowledge to come up with a three step stereo matching architecture and refined the parameters with the neural architecture search, \eg convolution filters sizes, strides, \etc. Other faster techniques use coarse-to-fine matching with hierarchical upsampling \cite{tankovich2021hitnet, hierarchical, deeppruner, pointstereo, sparsetodense, cascadecv}. Among them, DeepPruner achieves great performance. However, they ignore recovering details lost by lower resolution matching. Despite considerable research for the last half-century, a stereo matching architecture with a focus on practical implications is still an open question in the computer vision community, where a state-of-the-art algorithm can be applied to real-time applications. Our approach takes a step in this direction. 

%------------------------------------------------------------------------
\section{FRSNet: A Multi-scale Improvement}
\label{sec:methodology}
In this section, we describe the details of our proposed multi-scale iterative coarse-to-fine refinement framework for stereo matching (iCFR) and a proposed best performing network called FRSNet which uses this iCFR. The overall architecture is shown in \cref{fig:frsnet}.
%with the flow defining the iCFR.

\subsection{Feature Extraction}
As a feature extraction backbone, we employ a stacked hourglass network with skip connections between corresponding layers of the encoder and decoder \cite{ganet,aanet}. This densely connected network is shared by the input stereo image pair. We use the already calculated features at multiple scales from the decoder part (\ie $\frac{1}{3}, \frac{1}{6}, \frac{1}{12}, \frac{1}{24}, \frac{1}{48}$), as input to the prediction heads at the corresponding scales. \Cref{fig:frsnet} shows the feature extraction network with dense connections. These features are then either concatenated or correlated to form the cost volume which is then processed in the prediction head. 

\subsection{Prediction Head}
We combine the three latter parts of the stereo matching pipeline namely cost volume construction, cost aggregation and disparity regression into a \textit{prediction head}, for modularity. Modern volumetric end-to-end stereo matching architectures such as GC-Net \cite{gcnet} and GANet \cite{ganet} employ a 4D cost volume. This cost volume is built by either concatenating or correlating the features from the stereo image pair, extracted from the feature extraction backbone, typically a stacked hourglass network \cite{stackedhourglass} or a Feature Pyramid \cite{fpn}. Depending on the adaptation from any stereo matching network, the cost volume can have different dimensions. Typically a correlation cost volume requires much less memory and a lower number of learnable parameters. The cost volume is then filtered by the adapted strategy of the corresponding stereo matching architecture. Finally, the disparity regression layer estimates the output disparity map. Note that the prediction head is designed to be adopted from any stereo matching architecture. As shown in \cref{fig:predictionhead}, the prediction head can also produce multiple disparity maps from different stages of cost aggregation for intermediate supervision during training. The ground truth is down-sampled accordingly to calculate the loss for each of these predictions. \Cref{sec:gahead,sec:aahead} describe the working of the overall multi-scale architecture using prediction heads from two state-of-the-art networks GANet \cite{ganet} and AANet \cite{aanet}.

\subsubsection{Symmetric Cost Volume}
\begin{figure}
    \centering
    \includegraphics[width=0.8\columnwidth]{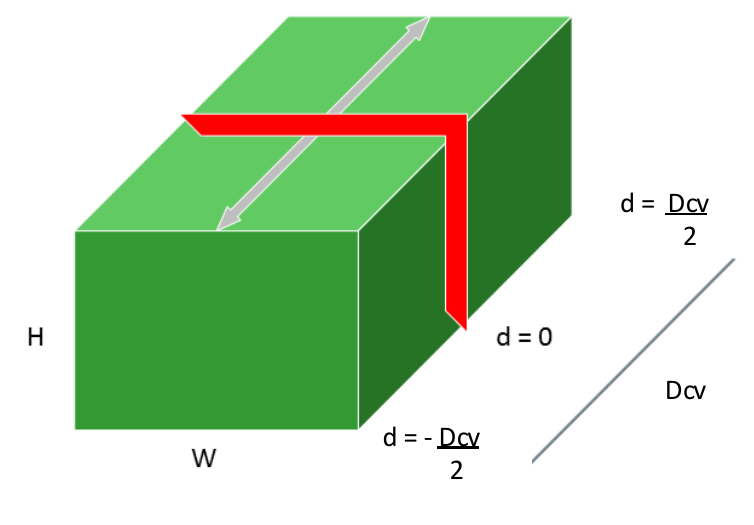}
    \caption{Illustration of the proposed symmetric cost volume to allow for negative disparity hypotheses. It ensures that initially too high estimates can be corrected by the residual disparities on finer scales.}
    \label{fig:symmetriccv}
\end{figure}

The disparity map produced at the smallest scale \ie $\frac{1}{48}$ is a very coarse estimate, usually consisting of just blobs of information. These blobs when up-sampled, produce large erroneous estimates of disparity initially. If these are not corrected in the next subsequent scale, the error keeps accumulating and amplifying much like a drift error. Negative disparity residuals are required to correct such coarse estimates. These residuals will help reduce the disparity values if they are initially too high. To this end, we change the way the cost volume is build to allow negative disparity hypotheses. \Cref{fig:symmetriccv} illustrates the construction of such a cost volume. The shape of the cost volume will remain the same $[F, D_{cv}, H, W]$, however the way it is filled along the disparity dimension, will be different. The idea is to fill the disparity dimension of length $D_{cv}$, with disparity hypotheses symmetrically in both directions, \ie positive and negative. This means that in position $\frac{D_{cv}}{2}$, the candidate with zero displacement will be placed. From here in both directions, the cost volume will be filled for the range $(\frac{-D_{cv}}{2}, \frac{D_{cv}}{2})$ respectively, leaving out one disparity candidate at the negative search space if required due to rounding. It is worth noting here that after this change the effective search space for disparity is $\frac{D_{cv}}{2}$.
%
%The importance of having such a cost volume in the proposed architecture is further demonstrated in the visualized predictions in Figure \ref{fig:neg_Disp_comp}.

\subsubsection{Disparity Regression}
For calculating the final disparity map we use disparity regression as proposed by Kendall \etal \cite{geometryandcontext}. This yields a much more continuous and consistent disparity map than just using classification based operations. The disparity prediction $\hat{d}$ is given by:
\begin{equation}
    \hat{d} = \sum^{\frac{D_{cv}}{2}}_{d=\frac{-D_{cv}}{2}} d \cdot \sigma (-C^{A}(d))
\end{equation}
Pseudo-probabilities for each disparity candidate $d$ are calculated from the filtered cost volume $C^{A}$ by applying the softmax operation $\sigma$. Note that the summation bounds in the equation above are consistent with the cost volume construction for negative disparity candidates.

\begin{algorithm}[b]
\caption{iCFR}
\label{alg:iterativeRefinement}
\begin{algorithmic}[1]
\State $[feat_{l}]$ = FeatureExtractor$(left)$%$res^{max}$
\State $[feat_{r}]$ = FeatureExtractor$(right)$
%\\
    \For{$(f_{l}, f_{r})$ in each scale}
            \If{smallest scale}
                \State $disp$ = PredictionHead$(f_{l}, f_{r})$
            \Else
                \State $disp$ = upSample$(disp)$
                \State $f_{r,warped}$ = Warp$(f_{r}, disp)$
                \State $residual$ = PredictionHead$(f_{l}, f_{r,warped})$
                \State $disp$ = $disp$ + $residual$
            \EndIf
    \EndFor
    \State $disp^{max}$ = Refinement$(disp, left, right)$
    \State \textbf{return} $disp^{max}$
\end{algorithmic}
\end{algorithm}
\subsection{iCFR: Iterative Coarse-to-Fine Refinement} \label{sec:IterativeRefinement}
As shown in \cref{fig:frsnet}, the overall network architecture works in a coarse-to-fine refinement fashion using features at multiple scales from the feature extractor. The initial coarse disparity map $disp^{i}$ is calculated at $\frac{1}{48}$ of the original resolution, with the prediction head for the corresponding scale. This $disp^{i}$ is then up-sampled to the next finer scale \ie $\frac{1}{24}$. Then the right image features at $\frac{1}{24}$th scale ($f_{r}^{i+1}$) are \textit{warped} towards this up-sampled $disp^{i'}$. The idea is that these warped features $f_{r,warped}^{i+1}$ will only represent the remaining disparity to be calculated. These $f_{r,warped}^{i+1}$ are then passed to the prediction head for $\frac{1}{24}th$ scale. Since the target image features are now warped, the prediction head will only produce a residual of the full disparity map $res^{i+1}$. Finally the up-sampled coarse disparity map at $disp^{i'}$ is added to this residual map $res^{i+1}$ to get a refined disparity map at this scale, \ie $disp^{i+1}$. This process is iterated for all scales \ie \{$\frac{1}{12}, \frac{1}{6}, \frac{1}{3}$\} to get the fine grained disparity map $disp^{i+4}$ at $\frac{1}{3}$ resolution. The $disp^{i+4}$ is then passed through the refinement network to get a final refined disparity map $disp^{max}$ at input resolution. A simplified pseudo-code of the iCFR is shown in \cref{alg:iterativeRefinement}.
Although the iterations at multiple resolutions introduce additional computational effort for calculating the final refined disparity, these computations follow a geometric series applying the sub-sampling factor (usually 2). This means the computational cost can only be twice \wrt the highest resolution for infinitely many sub-scales of iCFR. This is worth investing, considering the saved computations due to the smaller cost volumes at every scale.

% \begin{figure}[t]
% \minipage{0.5\columnwidth}
%   \includegraphics[width=\linewidth]{figures/no.nd.png}
% \endminipage\hfill
% \minipage{0.5\columnwidth}%
%   \includegraphics[width=\linewidth]{figures/nd.png}
% \endminipage
% \vspace{2mm}
% \caption[Negative Disparity Comparison]{Images show the results of having or not having a cost volume which can handle negative disparities. Left is the output without negative disparity handling and right is the output with negative disparity handling. The white regions indicate high negative values (color-mapped for visualization) which correspond to the high outlier rate in the experimental results in Tab \ref{sec:gahead}.}
% \label{fig:neg_Disp_comp}
% \end{figure}

\subsubsection{Disparity Search Range: $D_{cv}$}
Due to the use of prediction heads at different scales and because each subsequent disparity map is up-sampled for the next scale, the disparity values are also scaled appropriately. This means the $D_{cv}$ for the cost volume can be considerably reduced leading to a much shallower cost volume. For example a $D_{cv}$ of 12 at scale $\frac{1}{24}$ is scaled to 96 at the scale $\frac{1}{3}$. The $D_{cv}$ can be chosen so that at the highest scale, this value can estimate a $D_{max}$ of typically 192 pixels, which is usually seen in the state-of-the-art models. For any prediction head, the number of scales $S$ and the $D_{cv}$ is to be chosen correctly. The $D_{max}$ is given by:
\begin{equation}
\label{eqn:dmax}
D_{max} = \sum^{S}_{i=0} \frac{D_{cv}}{2} \cdot \frac{1}{s_{i}}
\end{equation}
where $s_{i}$ $\in$ \{$\frac{1}{48}, \frac{1}{24}, \frac{1}{12}, \frac{1}{6}, \frac{1}{3}$\} in the presented case.
The drastic effect of reduction in $D_{cv}$ on inference time is shown in \cref{tab:performance_comparison_kitti}.

\subsection{Refinement Network}
As discussed in \cref{sec:IterativeRefinement}, the $disp^{i+4}$ is produced at scale $\frac{1}{3}$. It typically has some edge gradient effects due to the multi-scale approach. These effects are especially visible on object boundaries and on areas where there is occlusion. To get rid of these artefacts, a final refinement block is applied. We use a stacked hourglass refinement layer as proposed by AANet+ \cite{aanet}. We calculate the photometric consistency error \cite{stereodrnet} and use that as an additional input to the refinement network, along with the left and right images. The refinement network then hierarchically up-samples the prediction at $\frac{1}{3}$ to input resolution and produces geometrically consistent and continuous disparity maps covering thin areas.
% \begin{table}
% \centering
% \caption{The table shows the reduction in inference time due to the change in $d_{max}$ value. }
% \label{tab:maxDisp_IT}
% \begin{tabular}{ |P{2cm}||P{2cm}|}
%  \hline
%  \multicolumn{2}{|c|}{FRSNet} \\
%  \hline\hline
%  Max Disp & Inference Time (secs) \\
%  \hline\hline
%  12 & 0.15\\
%  \hline
%  24 & 0.98\\
%  \hline
%  48 & 1.8\\
%  \hline \hline
%  192 & 7.2\\
%  \hline
% \end{tabular}
% \end{table}

% \begin{table*}
% \centering
% \caption{Evaluations of FRSNet with different design settings using the AAHead on Sceneflow dataset. Evaluation metrics are Average end point error (EPE) and 1 px threshold outlier rate (ER).}
% \label{tab:aaheadAblation}
% \resizebox{\textwidth}{!}{
% \begin{tabular}{P{3cm}|P{2cm}|P{3cm}|P{3cm}|P{2cm}|P{2cm}}
%  \hline\hline
%  \multicolumn{6}{c}{FRSNet Ablation experiments with AAHead - Sceneflow} \\
%  \hline
%  Multi-scale & Sym-CV & SDRNet \cite{sdrnet}-Ref & HG-Ref & EPE (px) & ER (\%) \\
%  \hline\hline
%  \checkmark & - & - & - & 4.91 & 43.1 \\
%  \hline
%  \checkmark & \checkmark & - & - & 2.08 & 23.2 \\
%  \hline
%  \checkmark & \checkmark & \checkmark & - & 1.90 & 21.4 \\
%  \hline
%  \checkmark & \checkmark  & - & \checkmark & \textbf{0.98} & \textbf{16.9} \\
%  \hline
% \end{tabular}}
% \end{table*}
\subsection{Loss Function}
The loss function that is typically used with the disparity regression layer is smooth $L1$ loss which we adopt in our pipeline too. This is because smooth $L1$ loss is quite robust against outliers and noise, and therefore helps against disparity discontinuities. We define our loss as:
\begin{equation}
    \hat{L}(\hat{d}, d) = \frac{1}{N} \sum_{i=1}^{N} L1(|\hat{d_{i}} - d_{i}|)
\end{equation}
\begin{equation}
    L1(x) = \begin{cases}
        x - 0.5, \quad &if \quad x \geq 1 \\
        x \cdot 0.5, \qquad  &if \quad x < 1    
    \end{cases}
\end{equation}
Here, $|\hat{d_{i}} - d_{i}|$ is the absolute error in the predicted disparities and $N$ is the number of ground truth label pixels.
The final loss function is the weighted sum of losses over all predictions:
\begin{equation}
\label{eqn:weightedloss}
    L = \sum_{i=1}^{M}\lambda_{i} \cdot {\Hat{L_{i}}}
\end{equation}
where $\lambda_{i}$ is a scalar weight and $M$ is the total number of predictions for intermediate supervision.

\begin{figure*}
    \centering
    \begin{subfigure}{\textwidth}
        \centering
        \includegraphics[width=\linewidth]{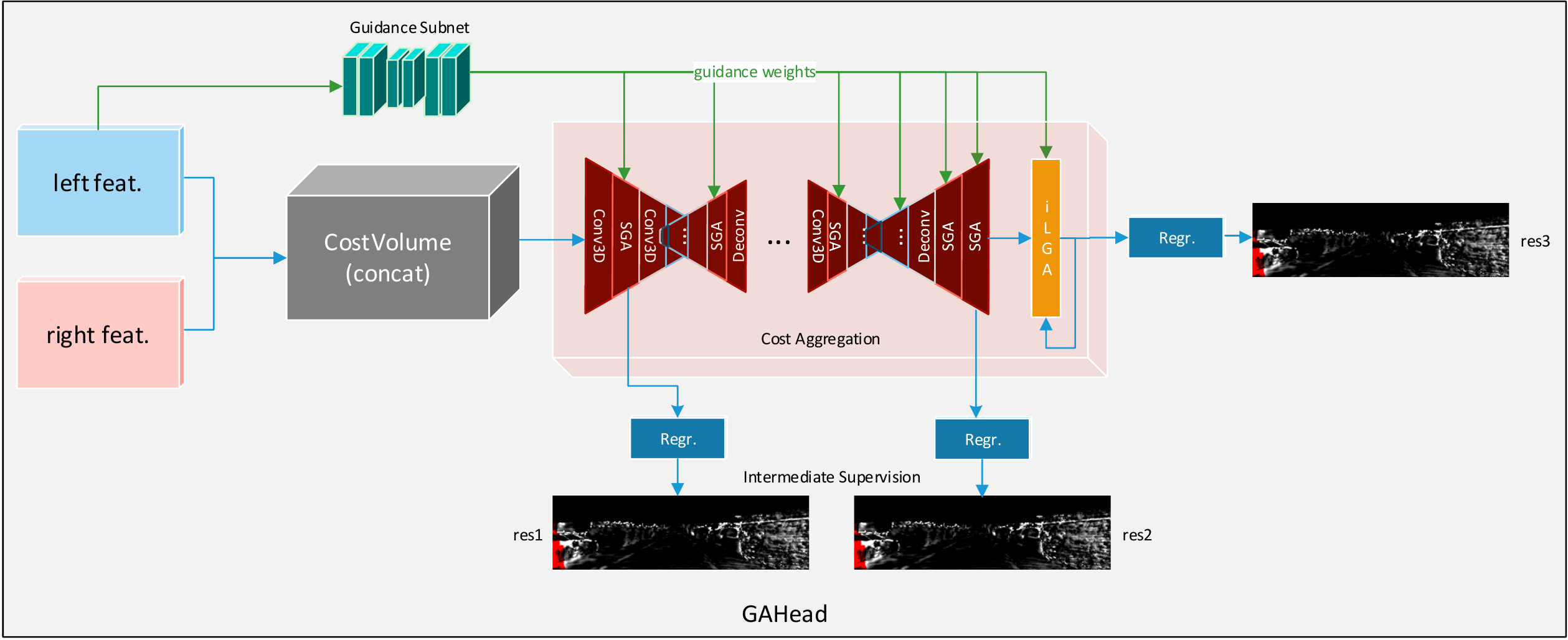}
        \caption{The prediction head adapted from GANet \cite{ganet}.}
        \label{fig:predictionhead:gahead}
    \end{subfigure}\\%
    \vspace{2mm}
    \begin{subfigure}{\textwidth}
        \centering
        \includegraphics[width=\linewidth]{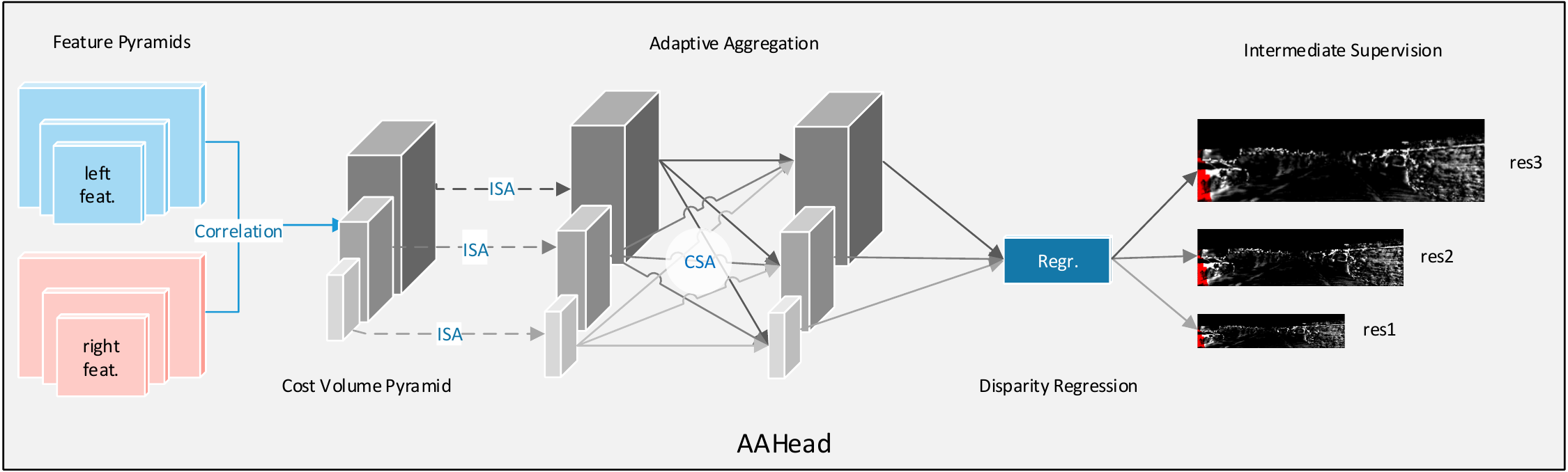}
        \caption{The prediction head adapted from AANet \cite{aanet}.}
        \label{fig:predictionhead:aahead}
    \end{subfigure}
    \caption{A more detailed view of the prediction heads as used in our iCFR. They can optionally contain a guidance subnet as in GANet \cite{ganet} (\subref{fig:predictionhead:gahead}) and support additional supervision with intermediate disparity regression. The inputs are the left and (warped) right image features at the corresponding scale from the feature extractor. These are used to build either a concatenation or correlation cost volume. The cost volume and the guidance weights are then passed to the cost aggregation or any kind of matching network. In our best performing network, ISA and CSA from AANet \cite{aanet} (\subref{fig:predictionhead:aahead}) are used for aggregation. Finally, a softmax regression layer estimates the final disparity or the residual disparity based on the refined cost volume.}
    \label{fig:predictionhead}
\end{figure*}

\subsection{GAHead}
\label{sec:gahead}
We apply the iCFR algorithm as described in \cref{sec:IterativeRefinement} on state-of-the-art GANet \cite{ganet} to significantly reduce its memory footprint and inference time. To this end we replace the prediction head in our network with the GANet matching network \ie the prediction head now consists of concatenation based 4D symmetric cost-volume, guidance subnet, cost aggregation network (SGA and LGA layers) and disparity regression network. We call it \textit{GAHead} and it is visualized in \cref{fig:predictionhead:gahead}. The guidance weights are calculated by the multi-scale guidance subnet on top of the feature extraction network which are reshaped and normalized for use in cost aggregation.
%This consists of several 2D convolutions for fast processing and is similar to what Zhang \etal \cite{ganet} proposed. It uses the left image features from the initial convolution of the feature extraction network. These guidance weights produced for each scale are used in SGA and LGA operations in the GAHead.
% \subsubsection{Multi-scale Guidance Subnet}
% The guidance weights are used in the cost aggregation network to guide the 3D convolutions towards the correct matching hypothesis as proposed by GANet \cite{ganet}. These weights if shared for all scales, will fail to capture the correct context for each scale and will thus produce discontinuities in the intermediate disparity residuals. We propose learning separate guidance weights for different scales so as to find the most precise matching candidate. Although this introduces additional learnable parameters, its positive effect is clearly seen on the EPE and ER, which we discuss in Section \ref{sec:experiments}.

\begin{table*}
\centering
\caption{Evaluations of FRSNet with different design settings using the GAHead and the AAHead on the Sceneflow dataset. Evaluation metrics are average end point error (EPE) and 1 px threshold outlier rate (ER).}
\label{tab:gaheadAblation}
\resizebox{\textwidth}{!}{
\begin{tabular}[t]{P{2cm}|P{2cm}|P{1.5cm}|P{1.5cm}|P{1.5cm}}
 \hline\hline
 \multicolumn{5}{c}{with GAHead}  \\
 %\hline
 Sym-CV & Improvements & HG-Ref & EPE [px] & ER [\%]\\
 \hline\hline
 - & - & - & 7.06 & 51.9 \\
 \hline
 \checkmark & - & - & 3.34 & 22.1  \\
 \hline
 \checkmark & \checkmark & - & 1.01 & 18.2 \\
 \hline
 \checkmark & \checkmark & \checkmark & \textbf{0.93} & \textbf{17.5} \\
 \hline
 \end{tabular}
\begin{tabular}[t]{P{2.5cm}|P{3cm}|P{1.5cm}|P{1.5cm}}
\hline\hline
 \multicolumn{4}{c}{with AAHead} \\
 %\hline
  Sym-CV & HG-Ref & EPE (px) & ER (\%) \\
  \hline\hline
  - & - & 4.91 & 43.1 \\
  \hline
  \checkmark & - & 2.08 & 23.2 \\
  \hline
  \checkmark  & \checkmark & \textbf{0.98} & \textbf{16.9}  \\
  %& & & \\
  \hline
\end{tabular}}
\end{table*}

\subsubsection{Improvements to the GAHead}
To push the GAHead performance even further we propose some improvements. Firstly, we replace a 3D convolution layer with an additional SGA layer, with additional guidance weights. This is because SGA layers are much faster than convolution layers and capture the disparity directly \cite{ganet, sgm}. Secondly, several LGA layers can be applied before and after passing it to the disparity regression layer to refine smaller details and fine structures in the final disparity map. However, instead of using multiple local guidance weights for each layer, we propose cost filtering using the same weights multiple times \ie in an iterative manner. This is inspired by the idea of CSPN \cite{cspn}. The filters are shared between iterations, thus keeping the number of learnable parameters constant. With each iteration, more details of the image are revealed which in turn improve the per-pixel depth estimation results. The number of iterations is experimentally determined with the best EPE obtained for $i=3$, without introducing too much overhead in run-time.
% \begin{table*}
% \centering
% \caption{Evaluations of FRSNet with different design dataset settings and different refinement modules on KITTI datasets. Evaluation metrics Average end point error (EPE) and D1 rate calculated on the KITTI 2012 and KITTI 2015 validation sets.}
% \label{tab:kittiAblation}
% \resizebox{\textwidth}{!}{
% \begin{tabular}{P{1.5cm}|P{2cm}|P{2cm}|P{1cm}|P{2cm}|P{2cm}|P{1cm}|P{1.5cm}|P{1.5cm}}
%  \hline\hline
%  \multicolumn{8}{c}{FRSNet fine-tuning experiments with KITTI 2012 and KITTI 2015} \\
%  \hline
%  Pred. Head & KITTI 2015 & KITTI 2012 & Mix & Pseudo-GT & SDRNet \cite{sdrnet}-Ref & HG-Ref & EPE (px) & D1 (\%) \\
%  \hline\hline
%  GAHead & \checkmark & - & - & - & - & - & 0.77 & 3.2 \\
%  \hline
%   & - & \checkmark & - & - & - & - & 0.80 & 3.6 \\
%  \hline
%   & \checkmark & - & - & - & \checkmark & - & 0.75 & \textbf{2.8} \\
%  \hline
%   & \checkmark & - & - & - &  & \checkmark & \textbf{0.74} & 2.9 \\
%  \hline
%  \hline
%  AAHead & \checkmark & - & - & - & - & - & 1.3 & 5.7 \\
%  \hline
%   & \checkmark & - & - & - & \checkmark & - & 0.78 & 2.9 \\
%  \hline
%   & \checkmark & - & - & - & - & \checkmark & 0.66 & 2.3 \\
%  \hline
%   & - & \checkmark & - & - & - & \checkmark & 0.66 & 2.7 \\
%  \hline
%   & \checkmark & \checkmark & \checkmark & - & - & \checkmark & 0.50 & 1.4 \\
%  \hline
%   & \checkmark & \checkmark & \checkmark & \checkmark & - & \checkmark & \textbf{0.47} & \textbf{1.0} \\
%  \hline
% \end{tabular}}
% \end{table*}

\subsection{AAHead}
\label{sec:aahead}
We also apply the iCFR algorithm on state-of-the-art AANet \cite{aanet} to reduce FLOPs and increase its scalability. To this end, we replace the prediction head with the adaptive aggregation module as described in AANet \cite{aanet}, which consists of intra/cross-scale aggregation operations. So the prediction head (AAHead, see \cref{fig:predictionhead:aahead}), now consists of a correlation-based 3D symmetric cost volume, an adaptive aggregation network and disparity regression. Since the AAHead works with feature pyramids, for each scale of the iCFR algorithm a nested feature pyramid at \{$\frac{1}{3}, \frac{1}{6}, \frac{1}{12}$\} sub-resolutions is constructed. The coarsest scale $\frac{1}{48}$ is omitted in this case because of the bottleneck in the adaptive aggregation module. To compensate for this reduction, the $D_{cv}$ is set to 24 according to \cref{eqn:dmax} to keep the $D_{max}$ close to 192. The adaptive aggregation network employs several modulated deformable convolution layers to aggregate the cost at different scales which has a higher receptive field than normal convolution operation \cite{deformableconvs}. Reduction in FLOPs due to lower resolutions and a much shallower cost volume results in a scalable and much more efficient overall model with even better sub-pixel accuracy and pixel outlier rate than the original AANet as shown in \cref{tab:performance_comparison_kitti}.

\begin{table*}
\centering
\caption{Comparison of the state-of-the-art GANet \cite{ganet} and AANet \cite{aanet} to the proposed FRSNet with corresponding prediction heads. Our proposed network is much more scalable in terms of GPU memory, FLOPs and the number of parameters. The inference time is also considerably reduced \ie $49\times$ faster than the GANet, making it real-time, without a drastic increase in the EPE and D1. Compared to the AANet \cite{aanet} it actually performs better by 0.05 px in EPE and 31\% better D1 on our validation set for KITTI 2015. All values are the mean results of running these models for an inference of 100 image pairs, on a GTX 1080Ti.
\label{tab:performance_comparison_kitti}
%The missing rows indicate that the model was unable to fit on the GPU with the corresponding input resolution.
For each model, the comparison stops at the highest resolution that fits into GPU memory.}
\resizebox{\textwidth}{!}{
\begin{tabular}{P{2cm}|P{2cm}|P{1cm}|P{1cm}|P{1.5cm}|P{1cm}|P{1.5cm}|P{1.5cm}|P{1.5cm}|P{1.5cm}}
 \hline\hline
%  \multicolumn{9}{c}{Efficiency Comparison - KITTI 2015} \\
 Method & Resolution & $D_{cv}$ & $D_{max}$ & Params & Mem & FLOPS & EPE [px] & D1 [\%] & Time [ms] \\
 \hline\hline
 GANet \cite{ganet} & KITTI & 48 & 192 & 6.5M & 6.2G & 2.2T & 0.54 & 1.80 & 7402 \\
%  & HD & 48 & 192 & 6.5M & - & - & - & - & -\\
%  & 4K & 48 & 192 & 6.5M & - & - & - & - & - \\
 \hline
 \multirow{3}{*}{FRSNet-GA} & KITTI & 4 & 186 & 6.5M & 1.4G & 373G & 0.74 & 2.80 & 150 \\
 & HD & 4 & 186 & 6.5M & 5.7G & 718G & - & - & 452 \\
 & 4K & 4 & 186 & 6.5M & 10G & 1.14T & - & - & 991 \\
 \hline\hline
 \multirow{2}{*}{AANet \cite{aanet}} & KITTI & 64 & 192 & 8.4M & 4.4G & 575G & 0.55 & 2.03 & 62\\
 & HD & 64 & 192 & 8.4M & 7.8G & 793G & - & - & 191\\
%  & 4K & 64 & 192 & 8.4M & - & - & - & - & -\\
 \hline
 \multirow{3}{*}{FRSNet-AA} & KITTI & 8 & 180 & 3.1M & 1.4G & 245G & 0.50 & 1.40 & 61\\
 & HD & 8 & 180 & 3.1M & 3.8G & 457G & - & - & 173\\
 & 4K & 8 & 180 & 3.1M & 9.9G & 981G & - & - & 692\\
 \hline
\end{tabular}
}
\end{table*}

\section{Experiments} \label{sec:experiments}
\begin{figure}
\centering
  \includegraphics[width=0.8\columnwidth]{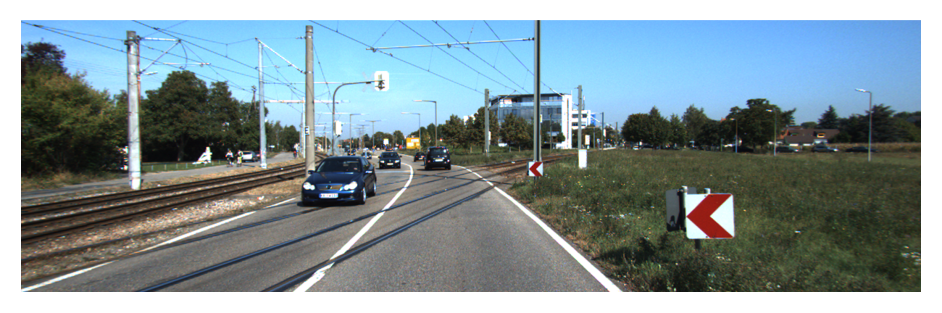}\\
  \includegraphics[width=0.8\columnwidth]{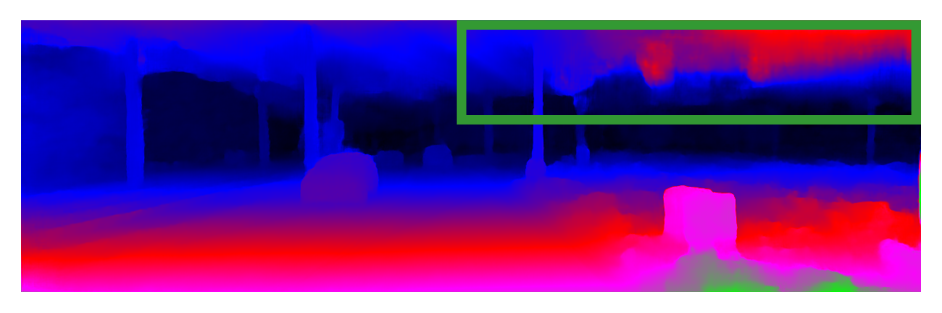}\\
  \includegraphics[width=0.8\columnwidth]{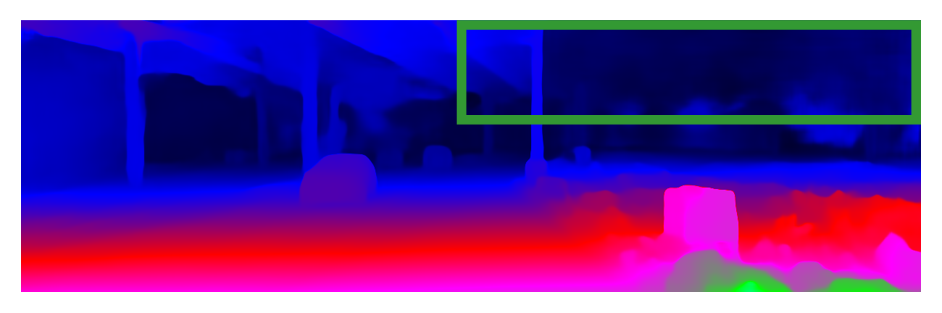}
\caption{Visual comparison of the results when training with (bottom) and without (middle) pseudo-GT. Using the pseudo-GT produces much more consistent results, \ie without extraneous disparity values on the areas where there is no GT especially in the sky regions. The quantitative effect of this training strategy is shown in \cref{tab:kittiAblation}.}
\label{fig:kitti_psuedo_GT}
\end{figure}

We perform exhaustive experiments on our FRSNet using the Sceneflow and KITTI 2015 datasets. The first proposed by Mayer \etal \cite{dispnet}, is a large dataset of synthetic images with dense ground truth. The KITTI dataset \cite{kitti} contains real-world outdoor images but with sparse ground truth. We report end-point-error (EPE) and 1-pixel outlier rate (ER) on the Sceneflow dataset and EPE and D1 rate on KITTI datasets. The network is implemented in PyTorch. Our final model uses Adam ($\beta_{1}=0.9, \beta_{2}=0.999$) as optimizer. 
We employ a three-stage training strategy for all our experiments. 1) Training on Sceneflow dataset, 2) Fine-tuning on KITTI dataset, 3) Optionally, fine-tuning on the (dense) pseudo-ground truth as proposed in AANet \cite{aanet}. This third step is required to produce visually consistent disparity maps for upper regions of KITTI images where there is no ground truth disparity available (\cf \cref{fig:kitti_psuedo_GT}).
%For details and visual proofs of improvement using this third stage, please refer to the supplementary material.
We set the $D_{cv}$ value to 12 and 24 for experiments with GAHead and AAHead, respectively. Training is done for 30 epochs with a batch size of 32 on two NVIDIA V100 32GB GPUs, with the crop-size of $240 \times 576$ on Sceneflow dataset and for 1000 epochs with a batch size of 16 on the KITTI datasets with the crop size of $384 \times 1248$. The third fine-tuning stage is carried out for a maximum of 8 epochs. The learning rate begins at 0.001 with a schedule for halving at 400th, 600th and 800th epoch in the fine-tuning phase. All the input image channels are normalized by subtracting their means and dividing their standard deviations. The loss weights $ \lambda_{i}$ in \cref{eqn:weightedloss} for multiple predictions are set to [$0.2, 0.4, 0.6, 1.0$] for the three (two intermediate + final) predictions of the finest pyramid scale and the ultimate prediction at input resolution of the refinement network, in this order.
% \subsection*{Pseudo Ground Truth}
% Finetuning on KITTI datasets for a large number of epochs reveal over-fitting artefacts in the top half of the predicted disparity maps. This is because the already sparse ground truth provided for the KITTI datasets have no disparity values in these top regions. The result is qualitatively worse and visually inconsistent disparity predictions. To get rid of these unwanted artefacts we use dense predictions produced by the pre-trained state-of-the-art GANet \cite{ganet} as ground truth for the third stage of fine-tuning as described in the start of Section \ref{sec:experiments} and similar to AANet \cite{aanet} for a few epochs. The result is visually consistent predictions as shown in Figure \ref{fig:kitti_psuedo_GT}.

\subsection{Ablation Study}
To validate the performance of the proposed iCFR algorithm on the proposed FRSNet, we perform several experiments on the Sceneflow test set and on our KITTI validation set of 15 image pairs, which are obtained by splitting the training set for KITTI 2015 dataset randomly. The ablation experiments involve using the GAHead and AAHead in FRSNet and evaluating the effectiveness of different improvements proposed in \cref{sec:methodology}. \Cref{tab:gaheadAblation} shows the overall results of the ablation study for both predictions heads on the Sceneflow test set. In both cases, it highlights the importance of having a symmetric cost volume (\textit{Sym-CV}) with negative disparity hypotheses which decreases the EPE and ER by almost half. With GAHead, the introduction of the proposed improvements lower the error metrics even further. It was noted that replacing a higher resolution 3D convolution layer in cost aggregation with a SGA layer not only reduces the EPE and ER but also reduces the inference time by 70 ms.
\par
\begin{figure}
\minipage{0.5\columnwidth}
  \includegraphics[width=\linewidth]{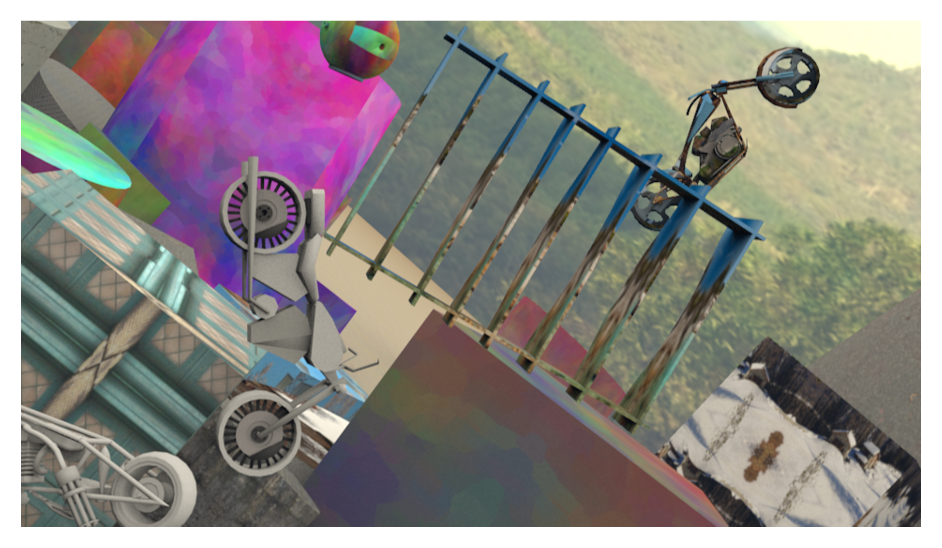}
\endminipage\hfill
\minipage{0.5\columnwidth}
  \includegraphics[width=\linewidth]{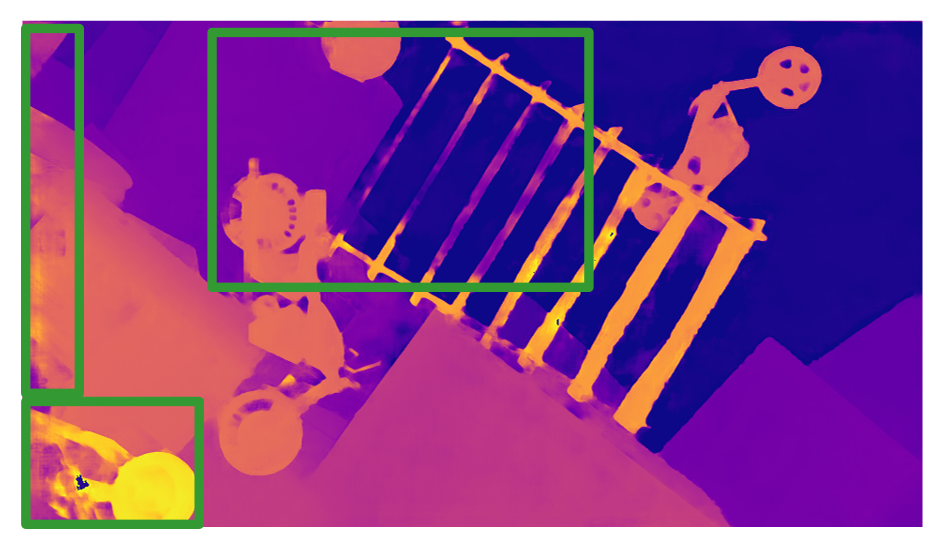}
\endminipage\hfill
\minipage{0.5\columnwidth}
  \includegraphics[width=\linewidth]{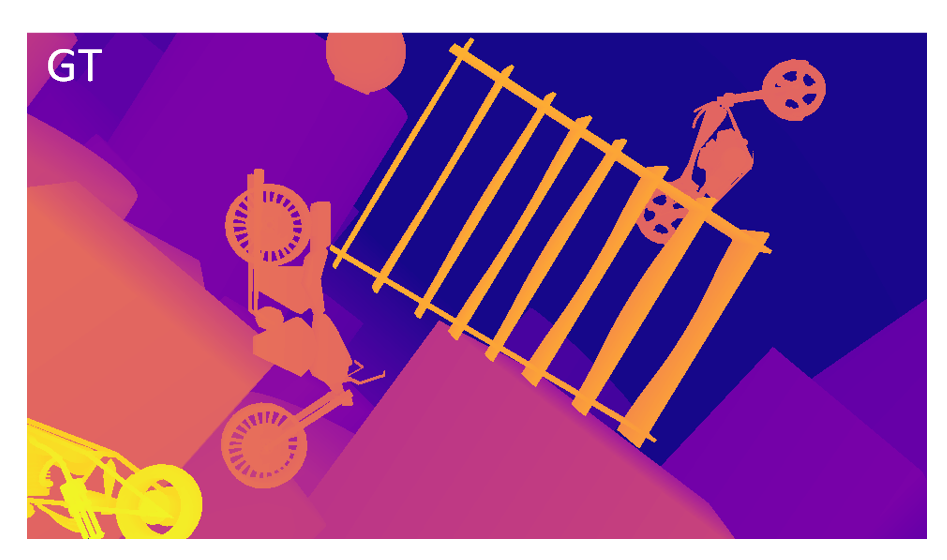}
\endminipage\hfill
\minipage{0.5\columnwidth}
  \includegraphics[width=\linewidth]{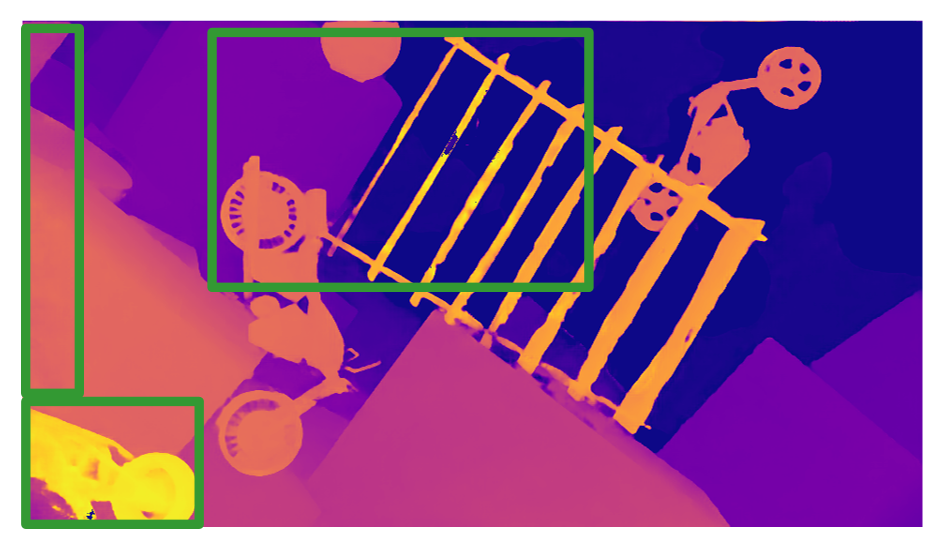}
\endminipage\hfill%
\minipage{\columnwidth}
\centering \includegraphics[width=0.99\columnwidth]{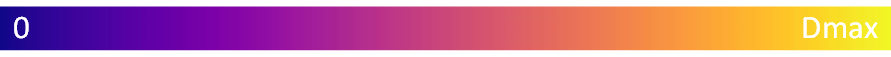}
\endminipage
\caption{Visual comparison on the Sceneflow test set, showing the impact of the refinement block. From top-left to bottom-right: Input reference images, prediction without refinement, GT disparity map, and refined prediction. Fine details are effectively recovered.}
\label{fig:refinement_comp}
\end{figure}

 Using the AAHead, without the final refinement layer yields relatively worse predictions than with GAHead.
%We experiment with the photometric consistency error refinement technique proposed in StereoDRNet \cite{stereodrnet} and achieve 2\% decrease in ER.
However, with a final hourglass refinement block (\textit{HG-Ref}) as proposed in AANet+ \cite{aanet} both variants of our model reach sub-pixel accuracy in EPE with the best ER. Similar improvements are seen while evaluating on the KITTI 2015 validation dataset. The positive impact of the refinement block is illustrated in \cref{fig:refinement_comp}. A quantitative evaluation is given in \Cref{tab:kittiAblation} in which the hourglass refinement block (\textit{HG-Ref}) is also compared against an alternative refinement module of SDRNet \cite{stereodrnet}. 

\subsection{Performance Comparison}
\begin{figure}
    \centering
    \includegraphics[width =\columnwidth]{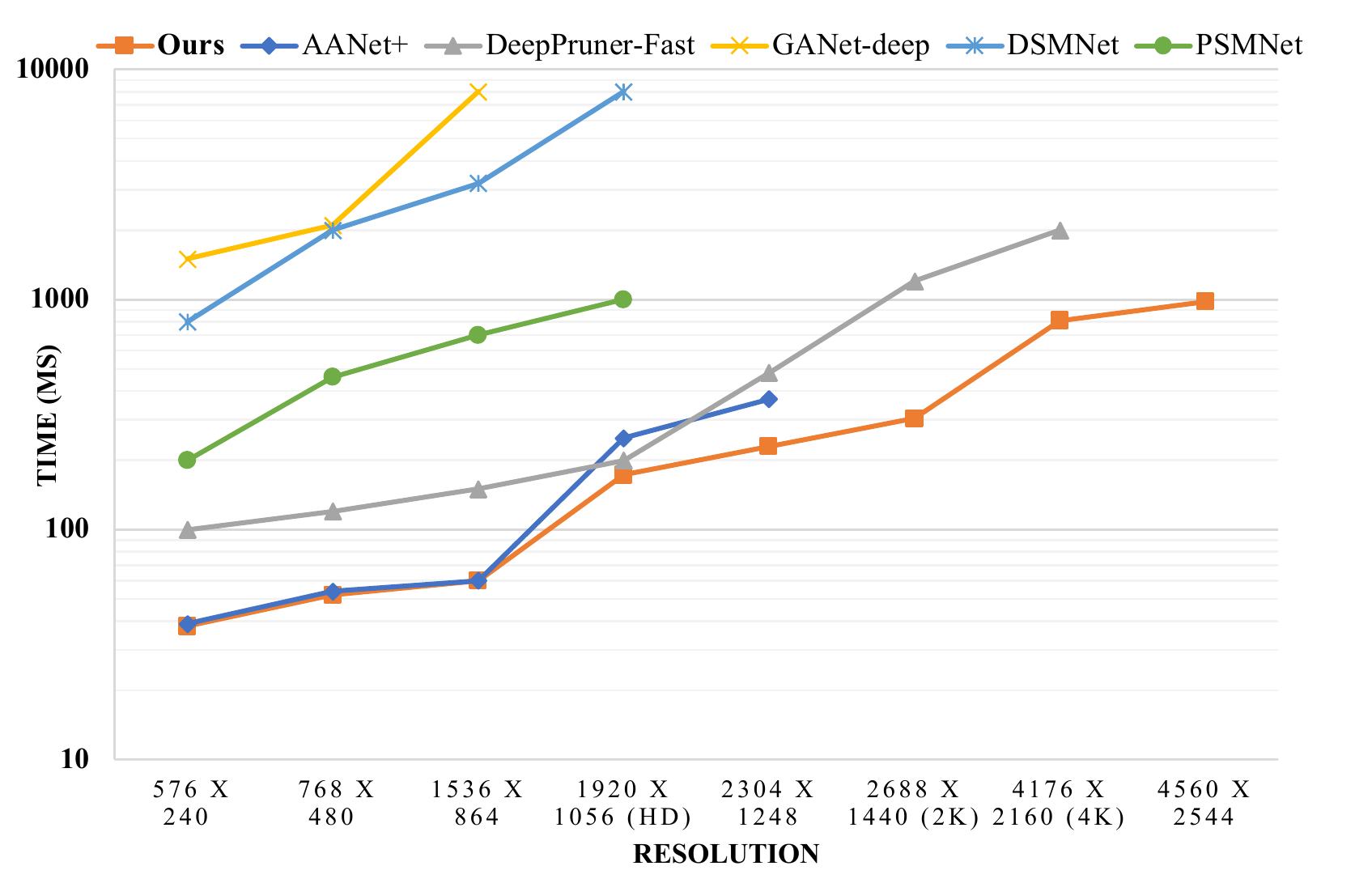}
    \caption{Run-time of different stereo matching models for varying input resolutions. For all measurements, a GTX 1080Ti GPU (12 GB) is used. Each curve ends where the corresponding model can no longer fit into the GPU memory. Our FRSNet is around 120$\times$ and 15$\times$ faster than the GANet \cite{ganet} and PSMNet \cite{psmnet}, respectively. AANet+ \cite{aanet} although having almost comparable run-time to ours, becomes too large to fit into the available GPU memory at just above HD resolution. DeepPruner \cite{deeppruner}, is 2.5$\times$ slower than our model, is less scalable in terms of memory footprint, and performs less accurately.}
    \label{fig:growth}
    %\vspace{-5mm}
\end{figure}

\Cref{tab:performance_comparison_kitti} shows a comprehensive comparison between our network with GAHead and AAHead and the corresponding original networks on our KITTI 2015 validation set. There is a significant decrease in the number of parameters, memory consumption, FLOPs and inference time. Our model only requires 16\% of the FLOPs required for the GANet-deep \cite{ganet}, while consuming 6$\times$ less memory. Similar improvement is seen compared to AANet \cite{aanet} where the memory requirement is less than half while D1 being 31\% better, with a comparable run-time. As also shown in \cref{fig:growth}, our model can scale up to even close to 6K resolution while keeping the run-time below one second. These performance improvements are made possible because of the lower $D_{cv}$ value, which in turn means a shallower cost volume. Moreover, the iterative computations at smaller resolutions are bounded by at most twice the time required for the common operations at the highest resolution. 

\begin{table}
\centering
\caption{Evaluation of our FRSNet for different training datasets and different refinement modules. Evaluation metrics EPE and D1 rate calculated on our KITTI 2015 validation set.}
\label{tab:kittiAblation}
\resizebox{\columnwidth}{!}{
\begin{tabular}{P{1cm}|P{1cm}|P{1cm}|P{1cm}|P{0.9cm}|P{1cm}|P{0.9cm}}
 \hline\hline
 KITTI 2015 & KITTI 2012 & Pseudo-GT & SDRNet \cite{stereodrnet}-Ref & HG-Ref & EPE (px) & D1 (\%) \\
 \hline\hline
 \checkmark & - & - & - & - & 1.3 & 5.7 \\
 \hline
 \checkmark & - & - & \checkmark & - & 0.78 & 2.9 \\
 \hline
 \checkmark & - & - & - & \checkmark & 0.66 & 2.3 \\
 \hline
  - & \checkmark & - & - & \checkmark & 0.66 & 2.7 \\
 \hline
 \checkmark & \checkmark & - &  \checkmark & - & 0.56 & 2.0 \\
 \hline
 \checkmark & \checkmark & \checkmark &  \checkmark & - & 0.54 & 1.7 \\
 \hline
 \checkmark & \checkmark & \checkmark & - & \checkmark & \textbf{0.50} & \textbf{1.4} \\
 \hline
\end{tabular}}
\end{table}

\begin{figure*}
\minipage{0.5\linewidth}
  \includegraphics[width=\linewidth]{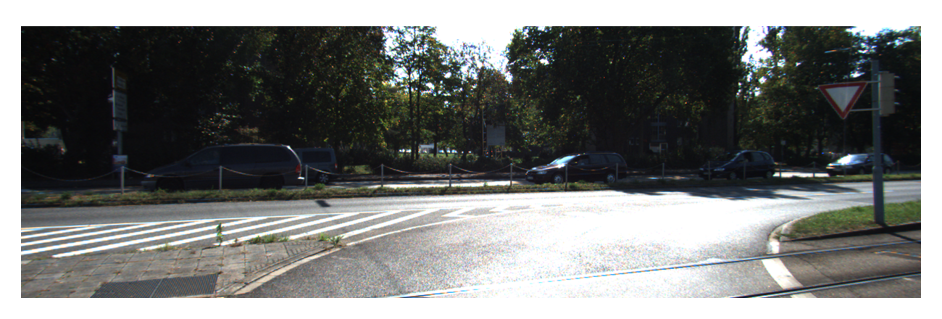}
\endminipage\hfill
\minipage{0.5\linewidth}
  \includegraphics[width=\linewidth]{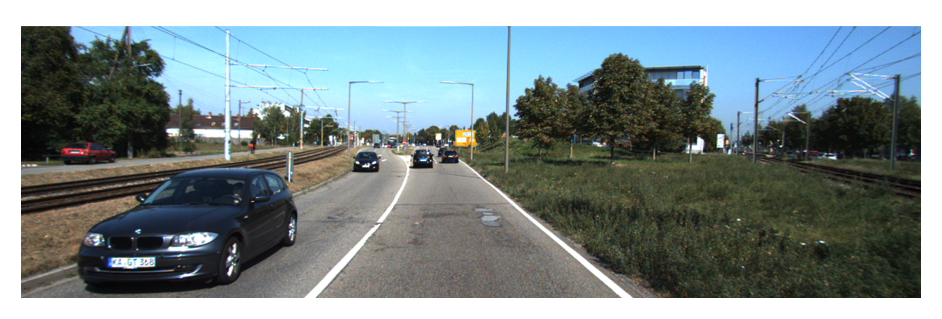}
\endminipage\hfill
\minipage{0.5\linewidth}
  \includegraphics[width=\linewidth]{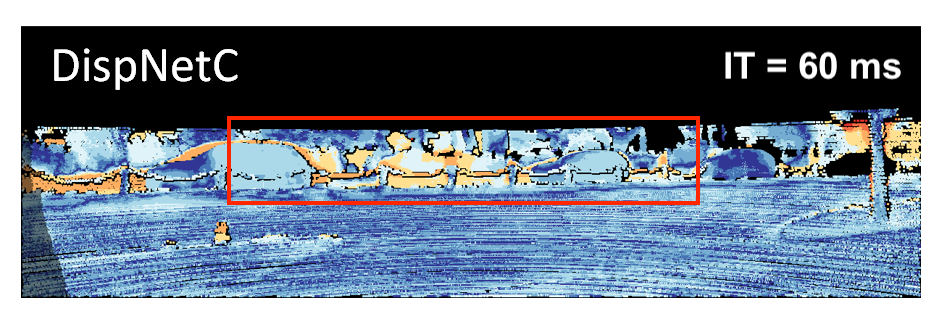}
\endminipage\hfill
\minipage{0.5\linewidth}
  \includegraphics[width=\linewidth]{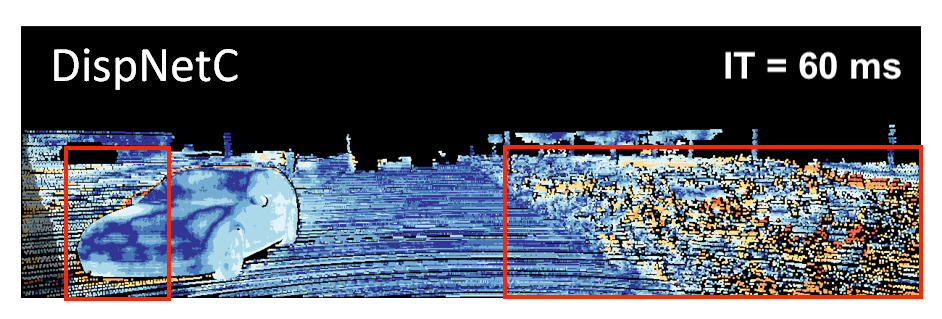}
\endminipage\hfill
\minipage{0.5\linewidth}
  \includegraphics[width=\linewidth]{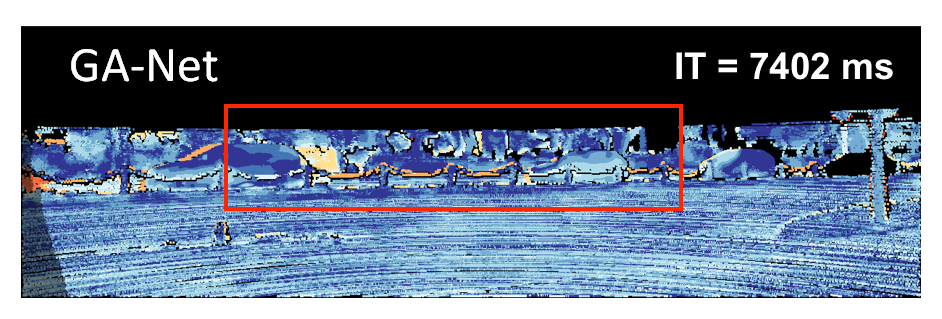}
\endminipage\hfill
\minipage{0.5\linewidth}%
  \includegraphics[width=\linewidth]{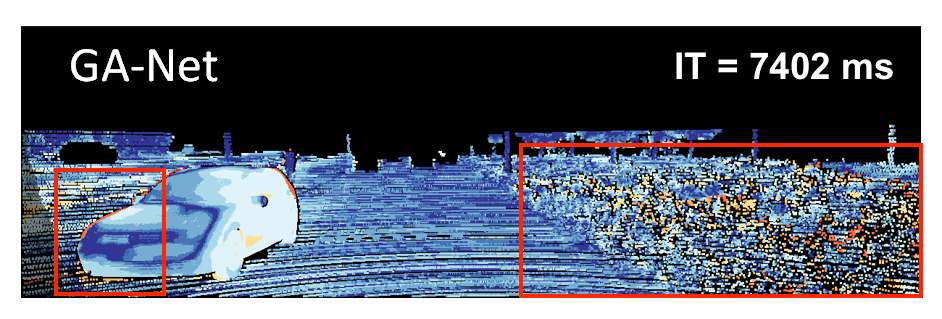}
\endminipage\hfill
\minipage{0.5\linewidth}%
  \includegraphics[width=\linewidth]{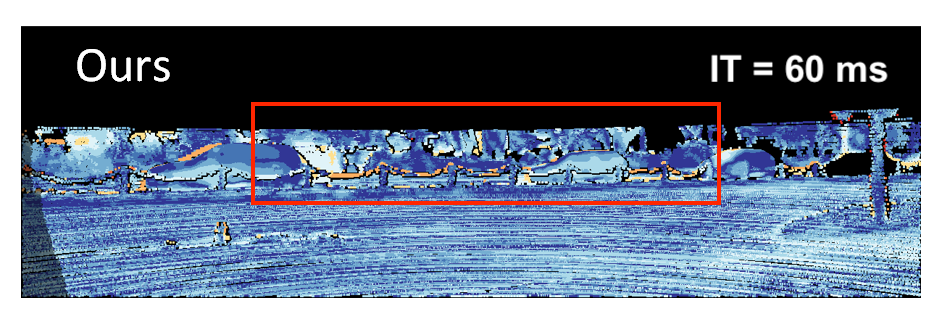}
\endminipage\hfill
\minipage{0.5\linewidth}%
  \includegraphics[width=\linewidth]{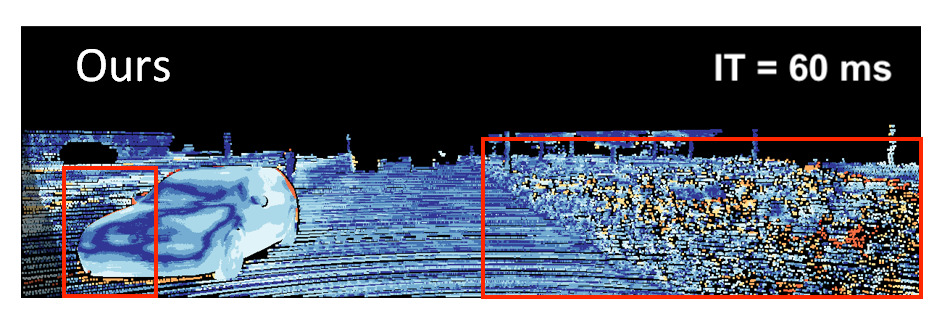}
\endminipage\hfill
\minipage{\linewidth}%
  \centering EPE: \includegraphics[width=0.9\linewidth]{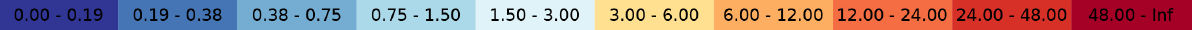}
\endminipage
\caption{Comparison of the D1 error maps on KITTI 2015 test set from top to bottom: DispNetC \cite{dispnet}, GANet \cite{ganet} and our FRSNet. Our model shows much smaller errors especially in occluded areas and at object boundaries as compared to the DispNetC. Compared to state-of-the-art GANet \cite{ganet} our model has even lower errors in smooth areas while being significantly more efficient.}
\label{fig:kitti_test_comp}
\end{figure*}

\subsection{Benchmark Results}
For benchmarking, we use our best performing network which is equipped with the AAHead and final hourglass refinement layer to evaluate on the KITTI 2015 test set on the official website. \Cref{tab:kittibenchmarkresult} shows the result of the evaluation in comparison to other models on the leader board. Compared to single-scale models our model achieves comparable results while reducing the complexity and run-time significantly. Compared to real-time models with inference time less than 100 milliseconds, our model achieves state-of-the-art results standing right beside the AANet+ (the improved variant of the original AANet) in D1 metric while having a lower memory and computational footprint and better scalability at the same inference time. The results in \Cref{tab:performance_comparison_sceneflow,tab:kittibenchmarkresult} for both the KITTI 2015 and Sceneflow datasets demonstrate that our method maintains a balance between accuracy and speed. The iCFR algorithm thus can be applied to any state-of-the-art stereo matching architecture to make it more efficient and scalable while retaining its accuracy.

\begin{table}
\centering
\caption{Benchmark results of the D1 metric (in \%) on the KITTI 2015 test set.
%All the values are displayed exactly as available on the online benchmark leader board.
Best numbers are given in bold, second-best are underlined. It is worth mentioning that the reported run-times are not measured in uniform settings, therefore a better comparison of run-time is provided by \cref{tab:performance_comparison_kitti} and \cref{fig:growth}.}
\label{tab:kittibenchmarkresult}
\resizebox{\columnwidth}{!}{
\begin{tabular}{c||c|c|c||c|c|c||c}
 \hline\hline
%  \multicolumn{8}{c}{Benchmark Results KITTI 2015} \\
%  \hline
 & \multicolumn{3}{c||}{D1-Noc} & \multicolumn{3}{c||}{D1-All} & Time\\
 %\hhline{~||---||---||}
 Model & bg & fg & all & bg & fg & all & [ms]\\
 \hline\hline
 GCNet \cite{gcnet}& 2.02 & \underline{3.12} & 2.45 & 2.21 & 6.16 &2.87 & \underline{900} \\
 PSMNet \cite{psmnet} & \underline{1.71} & 4.31 & \underline{2.14} & \underline{1.86} & \underline{4.62} & \underline{2.32} & \textbf{410} \\
 GANet \cite{ganet} & \textbf{1.34} & 3.11 & \textbf{1.63} & \textbf{1.48} & 3.46 & \textbf{1.81} & 1800 \\
 \hline
 StereoNet \cite{stereonet} & - & - & - & 4.30 & 7.45 & 4.83 & \textbf{15} \\
 DispNetC \cite{dispnet} & 4.11 & 3.72 & 4.05 & 4.32 & 4.41 & 4.34 & \underline{60} \\
 DeepPruner-fast \cite{deeppruner} & 2.13 & 3.43 & 2.35 & 2.32 & 3.91 & 2.59 & \underline{60} \\
 AANet+ \cite{aanet} & \textbf{1.49} & \underline{3.66} & \textbf{1.85} & \textbf{1.65} & \underline{3.96} & \textbf{2.03} & \underline{60} \\
 \textbf{FRSNet (Ours)} & \underline{1.58} & \textbf{3.45} & \underline{1.88} & \underline{1.73} & \textbf{3.87} & \underline{2.09} & \underline{60} \\
 \hline
 \end{tabular}
}
\end{table}

\begin{table}
\centering
\caption{Comparison of models on the Sceneflow dataset on our machine with a GTX 1080Ti GPU and using official code repositories. All the metrics are recorded after CUDA warm-up iterations, averaged on 100 inferences.}
\label{tab:performance_comparison_sceneflow}
\resizebox{\columnwidth}{!}{
\begin{tabular}{P{3cm}|P{1.5cm}|P{1.5cm}|P{1.5cm}}
 \hline\hline
%  \multicolumn{4}{c}{Performance Comparison - Sceneflow} \\
%  \hline
 Models & EPE [px] & $\geq$ 3px [\%] & Time [ms]\\
 \hline\hline
 GCNet \cite{gcnet} & 2.51 & \underline{9.34} & \underline{950} \\
 PSMNet \cite{psmnet} & \underline{1.09} & \textbf{4.14} & \textbf{640} \\
 GANet \cite{ganet} & \textbf{0.78} & - & 7402 \\
 \hline
 StereoNet \cite{stereonet} & 1.10 & - & \textbf{15} \\
 DispNetC \cite{dispnet} & 1.68 & 9.31 & \underline{60} \\
 DeepPruner \cite{deeppruner} & 0.97 & - & 120 \\
 AANet \cite{aanet} & \textbf{0.87} & \textbf{3.44} & 62\\
 \textbf{FRSNet (Ours)} & \underline{0.93} & \underline{6.01} & \underline{60} \\
\hline
\end{tabular}}
\end{table}

% \subsection*{Results Visualization}
% \par Figure \ref{fig:refinement_comp} show some output disparity maps of our network without the final hourglass refinement layer compared to the one with it. The former still produces continuous disparities in large regions but fail on smaller and thin objects. Moreover, object boundaries are much sharper in the refined predictions, thus reducing the D1 error as also shown in Table \ref{tab:gaheadAblation}. Figure \ref{fig:kitti_psuedo_GT} shows significant improvement in the visually consistent and qualitative predictions when the pseudo-GT is used for a small additional fine-tuning phase on KITTI datasets. This improves results significantly especially in areas where the GT disparity is missing \ie top part of the images.
\par \Cref{fig:kitti_test_comp} shows a comparison of D1 error maps produced by the KITTI 2015 benchmark on the test set. Our best performing model exhibits lower errors overall, especially at occluded areas and image boundaries. The error maps reveal even smaller errors on smooth regions. The maps also show that our model produces state-of-the-art results and outperforms other real-time models with visually consistent and continuous disparity maps.

\section{Conclusion}
In this paper, a novel multi-scale stereo matching architecture is proposed using the iCFR algorithm with disparity residuals. We show that our architecture can be applied to any state-of-the-art model to boost its efficiency in terms of run-time, memory, and scalability. We validate this on two state-of-the-art networks namely GANet \cite{ganet} and AANet \cite{aanet}. Our best performing model performs 120x faster than the GANet and uses 5x less memory, using just 16\% of FLOPs, while keeping the EPE and ER comparable. Our model scales until close to even 6K resolution on a GTX 1080Ti GPU, while keeping the inference time still below one second. The results show that by using iCFR, the trade-off between run-time and EPE can be optimized. Our best performing model produces state-of-the-art results on the Sceneflow test set as well as the KITTI 2015 benchmark while outperforming all the other real-time models in terms of accuracy.
\par Depending on the used prediction head and to keep the $D_{cv}$ low, the proposed architecture has a limitation on the number of scales (or the maximum down-sampling factor) to use. This is because of the relation between the cost aggregation technique, the $D_{max}$ and scales $s_{i}$ as shown in \cref{eqn:dmax}.
%To estimate the typical $D_{max}$ of 192, for example with the GAHead and $D_{cv}$ of 12, the required number of scales are five. Thus for a choice of a particular prediction head, the number of scales and the $D_{cv}$ need to be chosen correctly. However, these numbers can be calculated without much effort.
\par An interesting future work would be extending our architecture with already fast prediction heads to see how much further the performance improvement can be pushed. We also hope that the considerable reduction in the number of FLOPs will make our model suitable for edge computing devices with real-time performance requirements.

\begin{acks}
This work was partially funded by the Federal Ministry of Education and Research Germany under the project DECODE (01IW21001).
\end{acks}

\bibliographystyle{ACM-Reference-Format}
\bibliography{bib}


\begin{thebibliography}{31}

%%% the Web address.

\ifx \showCODEN    \undefined \def \showCODEN     #1{\unskip}     \fi
\ifx \showDOI      \undefined \def \showDOI       #1{#1}\fi
\ifx \showISBNx    \undefined \def \showISBNx     #1{\unskip}     \fi
\ifx \showISBNxiii \undefined \def \showISBNxiii  #1{\unskip}     \fi
\ifx \showISSN     \undefined \def \showISSN      #1{\unskip}     \fi
\ifx \showLCCN     \undefined \def \showLCCN      #1{\unskip}     \fi
\ifx \shownote     \undefined \def \shownote      #1{#1}          \fi
\ifx \showarticletitle \undefined \def \showarticletitle #1{#1}   \fi
\ifx \showURL      \undefined \def \showURL       {\relax}        \fi
% invisible to TeX
\providecommand\bibfield[2]{#2}
\providecommand\bibinfo[2]{#2}
\providecommand\natexlab[1]{#1}
\providecommand\showeprint[2][]{arXiv:#2}

\bibitem[\protect\citeauthoryear{Chabra, Straub, Sweeney, Newcombe, and
  Fuchs}{Chabra et~al\mbox{.}}{2019}]%
        {stereodrnet}
\bibfield{author}{\bibinfo{person}{Rohan Chabra}, \bibinfo{person}{Julian
  Straub}, \bibinfo{person}{Christopher Sweeney}, \bibinfo{person}{Richard
  Newcombe}, {and} \bibinfo{person}{Henry Fuchs}.}
  \bibinfo{year}{2019}\natexlab{}.
\newblock \showarticletitle{Stereodrnet: Dilated residual stereonet}. In
  \bibinfo{booktitle}{\emph{Conference on Computer Vision and Pattern
  Recognition (CVPR)}}.
\newblock


\bibitem[\protect\citeauthoryear{Chang and Chen}{Chang and Chen}{2018}]%
        {psmnet}
\bibfield{author}{\bibinfo{person}{Jia-Ren Chang} {and}
  \bibinfo{person}{Yong-Sheng Chen}.} \bibinfo{year}{2018}\natexlab{}.
\newblock \showarticletitle{Pyramid stereo matching network}. In
  \bibinfo{booktitle}{\emph{Conference on Computer Vision and Pattern
  Recognition (CVPR)}}.
\newblock


\bibitem[\protect\citeauthoryear{Chen, Han, Xu, and Su}{Chen
  et~al\mbox{.}}{2019}]%
        {pointstereo}
\bibfield{author}{\bibinfo{person}{Rui Chen}, \bibinfo{person}{Songfang Han},
  \bibinfo{person}{Jing Xu}, {and} \bibinfo{person}{Hao Su}.}
  \bibinfo{year}{2019}\natexlab{}.
\newblock \showarticletitle{Point-based multi-view stereo network}. In
  \bibinfo{booktitle}{\emph{International Conference on Computer Vision
  (ICCV)}}.
\newblock


\bibitem[\protect\citeauthoryear{Cheng, Wang, and Yang}{Cheng
  et~al\mbox{.}}{2019}]%
        {cspn}
\bibfield{author}{\bibinfo{person}{Xinjing Cheng}, \bibinfo{person}{Peng Wang},
  {and} \bibinfo{person}{Ruigang Yang}.} \bibinfo{year}{2019}\natexlab{}.
\newblock \showarticletitle{Learning depth with convolutional spatial
  propagation network}.
\newblock \bibinfo{journal}{\emph{Transactions on Pattern Analysis and Machine
  Intelligence (T-PAMI)}} (\bibinfo{year}{2019}).
\newblock


\bibitem[\protect\citeauthoryear{Cheng, Zhong, Harandi, Dai, Chang, Li,
  Drummond, and Ge}{Cheng et~al\mbox{.}}{2020}]%
        {leastereo}
\bibfield{author}{\bibinfo{person}{Xuelian Cheng}, \bibinfo{person}{Yiran
  Zhong}, \bibinfo{person}{Mehrtash Harandi}, \bibinfo{person}{Yuchao Dai},
  \bibinfo{person}{Xiaojun Chang}, \bibinfo{person}{Hongdong Li},
  \bibinfo{person}{Tom Drummond}, {and} \bibinfo{person}{Zongyuan Ge}.}
  \bibinfo{year}{2020}\natexlab{}.
\newblock \showarticletitle{Hierarchical Neural Architecture Search for Deep
  Stereo Matching}.
\newblock \bibinfo{journal}{\emph{Advances in Neural Information Processing
  Systems (NeurIPS)}} (\bibinfo{year}{2020}).
\newblock


\bibitem[\protect\citeauthoryear{Duggal, Wang, Ma, Hu, and Urtasun}{Duggal
  et~al\mbox{.}}{2019}]%
        {deeppruner}
\bibfield{author}{\bibinfo{person}{Shivam Duggal}, \bibinfo{person}{Shenlong
  Wang}, \bibinfo{person}{Wei-Chiu Ma}, \bibinfo{person}{Rui Hu}, {and}
  \bibinfo{person}{Raquel Urtasun}.} \bibinfo{year}{2019}\natexlab{}.
\newblock \showarticletitle{DeepPruner: Learning efficient stereo matching via
  differentiable patchmatch}. In \bibinfo{booktitle}{\emph{International
  Conference on Computer Vision (ICCV)}}.
\newblock


\bibitem[\protect\citeauthoryear{Falkenhagen}{Falkenhagen}{1997}]%
        {blockstereo}
\bibfield{author}{\bibinfo{person}{Lutz Falkenhagen}.}
  \bibinfo{year}{1997}\natexlab{}.
\newblock \showarticletitle{Hierarchical block-based disparity estimation
  considering neighbourhood constraints}. In
  \bibinfo{booktitle}{\emph{International Workshop on SNHC and 3D Imaging}}.
\newblock


\bibitem[\protect\citeauthoryear{Geiger, Lenz, and Urtasun}{Geiger
  et~al\mbox{.}}{2012}]%
        {kitti}
\bibfield{author}{\bibinfo{person}{Andreas Geiger}, \bibinfo{person}{Philip
  Lenz}, {and} \bibinfo{person}{Raquel Urtasun}.}
  \bibinfo{year}{2012}\natexlab{}.
\newblock \showarticletitle{Are we ready for autonomous driving? the kitti
  vision benchmark suite}. In \bibinfo{booktitle}{\emph{Conference on Computer
  Vision and Pattern Recognition (CVPR)}}.
\newblock


\bibitem[\protect\citeauthoryear{Gu, Fan, Zhu, Dai, Tan, and Tan}{Gu
  et~al\mbox{.}}{2020}]%
        {cascadecv}
\bibfield{author}{\bibinfo{person}{Xiaodong Gu}, \bibinfo{person}{Zhiwen Fan},
  \bibinfo{person}{Siyu Zhu}, \bibinfo{person}{Zuozhuo Dai},
  \bibinfo{person}{Feitong Tan}, {and} \bibinfo{person}{Ping Tan}.}
  \bibinfo{year}{2020}\natexlab{}.
\newblock \showarticletitle{Cascade cost volume for high-resolution multi-view
  stereo and stereo matching}. In \bibinfo{booktitle}{\emph{Conference on
  Computer Vision and Pattern Recognition (CVPR)}}.
\newblock


\bibitem[\protect\citeauthoryear{Hirschmuller}{Hirschmuller}{2005}]%
        {sgm}
\bibfield{author}{\bibinfo{person}{Heiko Hirschmuller}.}
  \bibinfo{year}{2005}\natexlab{}.
\newblock \showarticletitle{Accurate and efficient stereo processing by
  semi-global matching and mutual information}. In
  \bibinfo{booktitle}{\emph{Conference on Computer Vision and Pattern
  Recognition (CVPR)}}.
\newblock


\bibitem[\protect\citeauthoryear{Jancosek and Pajdla}{Jancosek and
  Pajdla}{2009}]%
        {segmentation}
\bibfield{author}{\bibinfo{person}{Michal Jancosek} {and}
  \bibinfo{person}{Tom{\'a}s Pajdla}.} \bibinfo{year}{2009}\natexlab{}.
\newblock \bibinfo{booktitle}{\emph{Segmentation based multi-view stereo}}.
\newblock \bibinfo{publisher}{Citeseer}.
\newblock


\bibitem[\protect\citeauthoryear{Kendall, Martirosyan, Dasgupta, Henry,
  Kennedy, Bachrach, and Bry}{Kendall et~al\mbox{.}}{2017a}]%
        {gcnet}
\bibfield{author}{\bibinfo{person}{Alex Kendall}, \bibinfo{person}{Hayk
  Martirosyan}, \bibinfo{person}{Saumitro Dasgupta}, \bibinfo{person}{Peter
  Henry}, \bibinfo{person}{Ryan Kennedy}, \bibinfo{person}{Abraham Bachrach},
  {and} \bibinfo{person}{Adam Bry}.} \bibinfo{year}{2017}\natexlab{a}.
\newblock \showarticletitle{End-to-end learning of geometry and context for
  deep stereo regression}. In \bibinfo{booktitle}{\emph{International
  Conference on Computer Vision (ICCV)}}.
\newblock


\bibitem[\protect\citeauthoryear{Kendall, Martirosyan, Dasgupta, Henry,
  Kennedy, Bachrach, and Bry}{Kendall et~al\mbox{.}}{2017b}]%
        {geometryandcontext}
\bibfield{author}{\bibinfo{person}{Alex Kendall}, \bibinfo{person}{Hayk
  Martirosyan}, \bibinfo{person}{Saumitro Dasgupta}, \bibinfo{person}{Peter
  Henry}, \bibinfo{person}{Ryan Kennedy}, \bibinfo{person}{Abraham Bachrach},
  {and} \bibinfo{person}{Adam Bry}.} \bibinfo{year}{2017}\natexlab{b}.
\newblock \showarticletitle{End-to-end learning of geometry and context for
  deep stereo regression}. In \bibinfo{booktitle}{\emph{International
  Conference on Computer Vision (ICCV)}}.
\newblock


\bibitem[\protect\citeauthoryear{Khamis, Fanello, Rhemann, Kowdle, Valentin,
  and Izadi}{Khamis et~al\mbox{.}}{2018}]%
        {stereonet}
\bibfield{author}{\bibinfo{person}{Sameh Khamis}, \bibinfo{person}{Sean
  Fanello}, \bibinfo{person}{Christoph Rhemann}, \bibinfo{person}{Adarsh
  Kowdle}, \bibinfo{person}{Julien Valentin}, {and} \bibinfo{person}{Shahram
  Izadi}.} \bibinfo{year}{2018}\natexlab{}.
\newblock \showarticletitle{StereoNet: Guided hierarchical refinement for
  real-time edge-aware depth prediction}. In \bibinfo{booktitle}{\emph{European
  Conference on Computer Vision (ECCV)}}.
\newblock


\bibitem[\protect\citeauthoryear{Lin, Doll{\'a}r, Girshick, He, Hariharan, and
  Belongie}{Lin et~al\mbox{.}}{2017}]%
        {fpn}
\bibfield{author}{\bibinfo{person}{Tsung-Yi Lin}, \bibinfo{person}{Piotr
  Doll{\'a}r}, \bibinfo{person}{Ross Girshick}, \bibinfo{person}{Kaiming He},
  \bibinfo{person}{Bharath Hariharan}, {and} \bibinfo{person}{Serge Belongie}.}
  \bibinfo{year}{2017}\natexlab{}.
\newblock \showarticletitle{Feature pyramid networks for object detection}. In
  \bibinfo{booktitle}{\emph{Conference on Computer Vision and Pattern
  Recognition (CVPR)}}.
\newblock


\bibitem[\protect\citeauthoryear{Liu, Cao, Dai, and Xu}{Liu
  et~al\mbox{.}}{2009}]%
        {continuouspatch}
\bibfield{author}{\bibinfo{person}{Yebin Liu}, \bibinfo{person}{Xun Cao},
  \bibinfo{person}{Qionghai Dai}, {and} \bibinfo{person}{Wenli Xu}.}
  \bibinfo{year}{2009}\natexlab{}.
\newblock \showarticletitle{Continuous depth estimation for multi-view stereo}.
  In \bibinfo{booktitle}{\emph{Conference on Computer Vision and Pattern
  Recognition (CVPR)}}.
\newblock


\bibitem[\protect\citeauthoryear{Mayer, Ilg, Hausser, Fischer, Cremers,
  Dosovitskiy, and Brox}{Mayer et~al\mbox{.}}{2016}]%
        {dispnet}
\bibfield{author}{\bibinfo{person}{Nikolaus Mayer}, \bibinfo{person}{Eddy Ilg},
  \bibinfo{person}{Philip Hausser}, \bibinfo{person}{Philipp Fischer},
  \bibinfo{person}{Daniel Cremers}, \bibinfo{person}{Alexey Dosovitskiy}, {and}
  \bibinfo{person}{Thomas Brox}.} \bibinfo{year}{2016}\natexlab{}.
\newblock \showarticletitle{A large dataset to train convolutional networks for
  disparity, optical flow, and scene flow estimation}. In
  \bibinfo{booktitle}{\emph{Conference on Computer Vision and Pattern
  Recognition (CVPR)}}.
\newblock


\bibitem[\protect\citeauthoryear{Min, Lu, and Do}{Min et~al\mbox{.}}{2011}]%
        {compredundancy}
\bibfield{author}{\bibinfo{person}{Dongbo Min}, \bibinfo{person}{Jiangbo Lu},
  {and} \bibinfo{person}{Minh~N Do}.} \bibinfo{year}{2011}\natexlab{}.
\newblock \showarticletitle{A revisit to cost aggregation in stereo matching:
  How far can we reduce its computational redundancy?}. In
  \bibinfo{booktitle}{\emph{International Conference on Computer Vision
  (ICCV)}}.
\newblock


\bibitem[\protect\citeauthoryear{Newell, Yang, and Deng}{Newell
  et~al\mbox{.}}{2016}]%
        {stackedhourglass}
\bibfield{author}{\bibinfo{person}{Alejandro Newell}, \bibinfo{person}{Kaiyu
  Yang}, {and} \bibinfo{person}{Jia Deng}.} \bibinfo{year}{2016}\natexlab{}.
\newblock \showarticletitle{Stacked hourglass networks for human pose
  estimation}. In \bibinfo{booktitle}{\emph{European Conference on Computer
  Vision (ECCV)}}.
\newblock


\bibitem[\protect\citeauthoryear{Pang, Sun, Ren, Yang, and Yan}{Pang
  et~al\mbox{.}}{2017}]%
        {cascaderesidual}
\bibfield{author}{\bibinfo{person}{Jiahao Pang}, \bibinfo{person}{Wenxiu Sun},
  \bibinfo{person}{Jimmy~SJ Ren}, \bibinfo{person}{Chengxi Yang}, {and}
  \bibinfo{person}{Qiong Yan}.} \bibinfo{year}{2017}\natexlab{}.
\newblock \showarticletitle{Cascade residual learning: A two-stage
  convolutional neural network for stereo matching}. In
  \bibinfo{booktitle}{\emph{International Conference on Computer Vision
  Workshops (ICCVW)}}.
\newblock


\bibitem[\protect\citeauthoryear{Schonberger, Sinha, and Pollefeys}{Schonberger
  et~al\mbox{.}}{2018}]%
        {fuseproposals}
\bibfield{author}{\bibinfo{person}{Johannes~L Schonberger},
  \bibinfo{person}{Sudipta~N Sinha}, {and} \bibinfo{person}{Marc Pollefeys}.}
  \bibinfo{year}{2018}\natexlab{}.
\newblock \showarticletitle{Learning to fuse proposals from multiple scanline
  optimizations in semi-global matching}. In \bibinfo{booktitle}{\emph{European
  Conference on Computer Vision (ECCV)}}.
\newblock


\bibitem[\protect\citeauthoryear{Tankovich, Hane, Zhang, Kowdle, Fanello, and
  Bouaziz}{Tankovich et~al\mbox{.}}{2021}]%
        {tankovich2021hitnet}
\bibfield{author}{\bibinfo{person}{Vladimir Tankovich},
  \bibinfo{person}{Christian Hane}, \bibinfo{person}{Yinda Zhang},
  \bibinfo{person}{Adarsh Kowdle}, \bibinfo{person}{Sean Fanello}, {and}
  \bibinfo{person}{Sofien Bouaziz}.} \bibinfo{year}{2021}\natexlab{}.
\newblock \showarticletitle{HITNet: Hierarchical iterative tile refinement
  network for real-time stereo matching}. In
  \bibinfo{booktitle}{\emph{Conference on Computer Vision and Pattern
  Recognition (CVPR)}}.
\newblock


\bibitem[\protect\citeauthoryear{Xu and Zhang}{Xu and Zhang}{2020}]%
        {aanet}
\bibfield{author}{\bibinfo{person}{Haofei Xu} {and} \bibinfo{person}{Juyong
  Zhang}.} \bibinfo{year}{2020}\natexlab{}.
\newblock \showarticletitle{AANet: Adaptive aggregation network for efficient
  stereo matching}. In \bibinfo{booktitle}{\emph{Conference on Computer Vision
  and Pattern Recognition (CVPR)}}.
\newblock


\bibitem[\protect\citeauthoryear{Yang}{Yang}{2012}]%
        {nonlocalcost}
\bibfield{author}{\bibinfo{person}{Qingxiong Yang}.}
  \bibinfo{year}{2012}\natexlab{}.
\newblock \showarticletitle{A non-local cost aggregation method for stereo
  matching}. In \bibinfo{booktitle}{\emph{Conference on Computer Vision and
  Pattern Recognition (CVPR)}}.
\newblock


\bibitem[\protect\citeauthoryear{Yin, Darrell, and Yu}{Yin
  et~al\mbox{.}}{2019}]%
        {hierarchical}
\bibfield{author}{\bibinfo{person}{Zhichao Yin}, \bibinfo{person}{Trevor
  Darrell}, {and} \bibinfo{person}{Fisher Yu}.}
  \bibinfo{year}{2019}\natexlab{}.
\newblock \showarticletitle{Hierarchical discrete distribution decomposition
  for match density estimation}. In \bibinfo{booktitle}{\emph{Conference on
  Computer Vision and Pattern Recognition (CVPR)}}.
\newblock


\bibitem[\protect\citeauthoryear{Yu and Gao}{Yu and Gao}{2020}]%
        {sparsetodense}
\bibfield{author}{\bibinfo{person}{Zehao Yu} {and} \bibinfo{person}{Shenghua
  Gao}.} \bibinfo{year}{2020}\natexlab{}.
\newblock \showarticletitle{Fast-MVSnet: Sparse-to-dense multi-view stereo with
  learned propagation and gauss-newton refinement}. In
  \bibinfo{booktitle}{\emph{Conference on Computer Vision and Pattern
  Recognition (CVPR)}}.
\newblock


\bibitem[\protect\citeauthoryear{Zbontar and LeCun}{Zbontar and LeCun}{2015a}]%
        {toast}
\bibfield{author}{\bibinfo{person}{Jure Zbontar} {and} \bibinfo{person}{Yann
  LeCun}.} \bibinfo{year}{2015}\natexlab{a}.
\newblock \showarticletitle{Computing the stereo matching cost with a
  convolutional neural network}. In \bibinfo{booktitle}{\emph{Conference on
  Computer Vision and Pattern Recognition (CVPR)}}.
\newblock


\bibitem[\protect\citeauthoryear{Zbontar and LeCun}{Zbontar and LeCun}{2015b}]%
        {mccnn}
\bibfield{author}{\bibinfo{person}{Jure Zbontar} {and} \bibinfo{person}{Yann
  LeCun}.} \bibinfo{year}{2015}\natexlab{b}.
\newblock \showarticletitle{Computing the stereo matching cost with a
  convolutional neural network}. In \bibinfo{booktitle}{\emph{Conference on
  Computer Vision and Pattern Recognition (CVPR)}}.
\newblock


\bibitem[\protect\citeauthoryear{Zhang, Prisacariu, Yang, and Torr}{Zhang
  et~al\mbox{.}}{2019}]%
        {ganet}
\bibfield{author}{\bibinfo{person}{Feihu Zhang}, \bibinfo{person}{Victor
  Prisacariu}, \bibinfo{person}{Ruigang Yang}, {and} \bibinfo{person}{Philip~HS
  Torr}.} \bibinfo{year}{2019}\natexlab{}.
\newblock \showarticletitle{GA-Net: Guided aggregation net for end-to-end
  stereo matching}. In \bibinfo{booktitle}{\emph{Conference on Computer Vision
  and Pattern Recognition (CVPR)}}.
\newblock


\bibitem[\protect\citeauthoryear{Zhang and Liu}{Zhang and Liu}{2014}]%
        {survey}
\bibfield{author}{\bibinfo{person}{Xiaoxue Zhang} {and}
  \bibinfo{person}{Zhigang Liu}.} \bibinfo{year}{2014}\natexlab{}.
\newblock \showarticletitle{A survey on stereo vision matching algorithms}. In
  \bibinfo{booktitle}{\emph{World Congress on Intelligent Control and
  Automation (WCICA)}}.
\newblock


\bibitem[\protect\citeauthoryear{Zhu, Hu, Lin, and Dai}{Zhu
  et~al\mbox{.}}{2019}]%
        {deformableconvs}
\bibfield{author}{\bibinfo{person}{Xizhou Zhu}, \bibinfo{person}{Han Hu},
  \bibinfo{person}{Stephen Lin}, {and} \bibinfo{person}{Jifeng Dai}.}
  \bibinfo{year}{2019}\natexlab{}.
\newblock \showarticletitle{Deformable convnets v2: More deformable, better
  results}. In \bibinfo{booktitle}{\emph{Conference on Computer Vision and
  Pattern Recognition (CVPR)}}.
\newblock


\end{thebibliography}

\end{document}